\documentclass[10pt,journal,draftcls,letterpaper,onecolumn]{IEEEtran}
\usepackage{times}
\usepackage{epsfig}
\usepackage{epstopdf}
\usepackage{graphicx}
\usepackage{amsmath}
\usepackage{amssymb}
\usepackage{amsthm}
\usepackage{cite}
\usepackage{booktabs}
\usepackage{algorithmic}
\usepackage{algorithm}
\usepackage{tabularx}
\usepackage{longtable}
\usepackage{bm}
\usepackage{color}
\usepackage[pagebackref=true,breaklinks=true,letterpaper=true,colorlinks,citecolor=blue,linkcolor=blue,bookmarks=false]{hyperref}

\usepackage{url}
\usepackage{caption}
\newcommand{\ignore}[1]{}

\usepackage{graphicx}
\usepackage{amsmath,amssymb} 
\usepackage{color}

\ifCLASSOPTIONcompsoc
\else
\fi

\begin{document}
%
\title{ Fast Tracking via Spatio-Temporal \\ Context Learning}

\author{{\quad\\}Kaihua~Zhang, Lei~Zhang, 
        Ming-Hsuan~Yang, and David~Zhang
\IEEEcompsocitemizethanks{
\IEEEcompsocthanksitem Kaihua~Zhang, Lei~Zhang and David~Zhang are with the Department of Computing, the Hong Kong Polytechnic University, Hong Kong.
E-mail: \{cskhzhang,cslzhang,csdzhang\}@comp.polyu.edu.hk.
\IEEEcompsocthanksitem Ming-Hsuan Yang is with Electrical Engineering and Computer Science, University of California, Merced, CA, 95344. E-mail: mhyang@ucmerced.edu.}
\thanks{}}

\markboth{}%
{Shell \MakeLowercase{\textit{et al.}}: Bare Demo of IEEEtran.cls for Computer Society Journals}
\IEEEcompsoctitleabstractindextext{%
\begin{abstract}
In this paper, we present a simple yet fast and robust
algorithm which exploits the spatio-temporal context for visual
tracking.
Our approach formulates the spatio-temporal relationships
between the object of interest and its local context based on a
Bayesian framework, which models the statistical
correlation between the low-level features (i.e., image intensity and
position) from the target and its surrounding regions.
The tracking problem is
posed by computing a confidence map, and obtaining the best
target location by maximizing an object location likelihood
function.
The Fast Fourier Transform is adopted for fast
learning and detection in this work.
Implemented in MATLAB without code optimization,
the proposed tracker runs at 350 frames per second on an i7
machine.
Extensive experimental results show that the proposed algorithm
performs favorably against state-of-the-art methods in terms of
efficiency, accuracy and robustness.
\end{abstract}

}

\maketitle

\IEEEdisplaynotcompsoctitleabstractindextext

%
\IEEEpeerreviewmaketitle

\newpage
\section{Introduction}

Visual tracking is one of the most active research topics due to its
wide range of applications such as motion analysis, activity
recognition, surveillance, and human-computer
interaction, to name a few~\cite{Yilmaz_CSUR_2006}.
The main challenge for robust visual
tracking is to handle large appearance changes of the target
object and the background over time due to occlusion,
illumination changes, and pose variation.
Numerous algorithms have been proposed with focus on effective
appearance models, which can be
categorized into
generative~\cite{collins2003mean,Collins_PAMI_2005,
  Adam_CVPR_2006, yang2007spatial, Ross_IJCV_2008,
  Kwon_CVPR_2010,kwon2011tracking,
  Mei_PAMI_2011,Li_CVPR_2011, sevilla2012distribution,
  oron2012locally, zhang2012robust} and
discriminative~\cite{Grabner_BMVC_2006,Kalal_CVPR_2010, Hare_ICCV_2011,
  Babenko_PAMI_2011, Zhang_ECCV_2012,
  henriques2012circulant} approaches.

\begin{figure}[t]
\begin{center}
\includegraphics[width=0.75\linewidth]{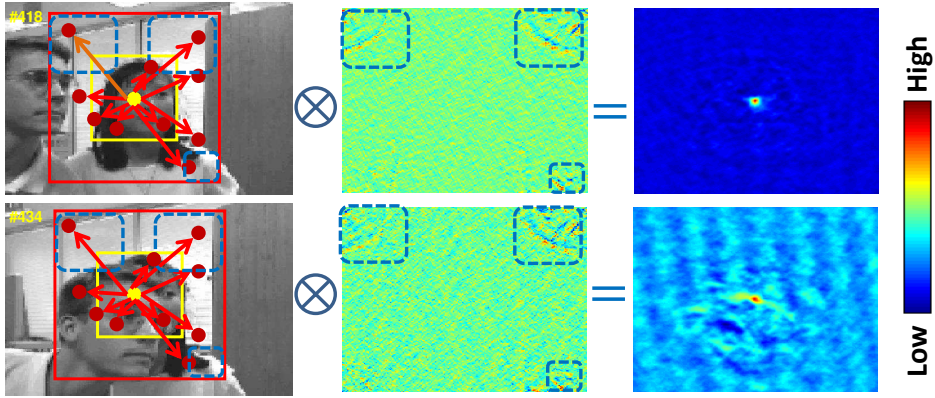}
\vspace{-2mm}
\end{center}
\caption{
The proposed method handles heavy occlusion well by learning spatio-temporal
context information.
Note that the region inside the
    red rectangle is the context region which includes the target and
    its surrounding background.
Left: although the target appearance
  changes much due to heavy occlusion, the spatial relationship
  between the object center (denoted by solid yellow circle) and its
  surrounding locations in the context region (denoted by solid red
  circles) is almost unchanged.
Middle: the learned spatio-temporal context
  model (the regions inside the blue rectangles have similar values
  which show the corresponding regions in the left frames have similar
  spatial relations to the target center.).
Right: the learned confidence map.}
\label{fig:demoocc}
\end{figure}

A generative tracking method learns an appearance model to represent
the target and search for image regions with best matching scores as
the results.
While it is critical to construct an effective
appearance model in order to to handle various challenging
factors in tracking,
the involved computational complexity is often increased at the
same time.
Furthermore, generative methods discard useful
information surrounding target regions that can be exploited to
better separate objects from backgrounds.
Discriminative methods treat
tracking as a binary classification problem with local search
which estimates decision boundary between an object image patch and
the background.
However, the objective of classification is to predict instance
labels
which is different from the goal of tracking
to estimate object locations~\cite{Hare_ICCV_2011}.
Moreover, while some efficient feature extraction techniques (e.g.,
integral image~\cite{Grabner_BMVC_2006,Kalal_CVPR_2010,Hare_ICCV_2011,
  Babenko_PAMI_2011,Zhang_ECCV_2012} and
random projection~\cite{Zhang_ECCV_2012}) have been proposed for
visual tracking, there often exist a large number of samples from which
features need to be extracted for classification, thereby
entailing computationally expensive operations.
Generally speaking, both generative and discriminative tracking
algorithms make trade-offs between effectiveness and efficiency of
an appearance model.
Notwithstanding much progress has been made in recent years, it
remains a challenging task to develop an efficient and robust tracking
algorithm.

\begin{figure*}[t]
\begin{center}
\includegraphics[width=0.85\linewidth]{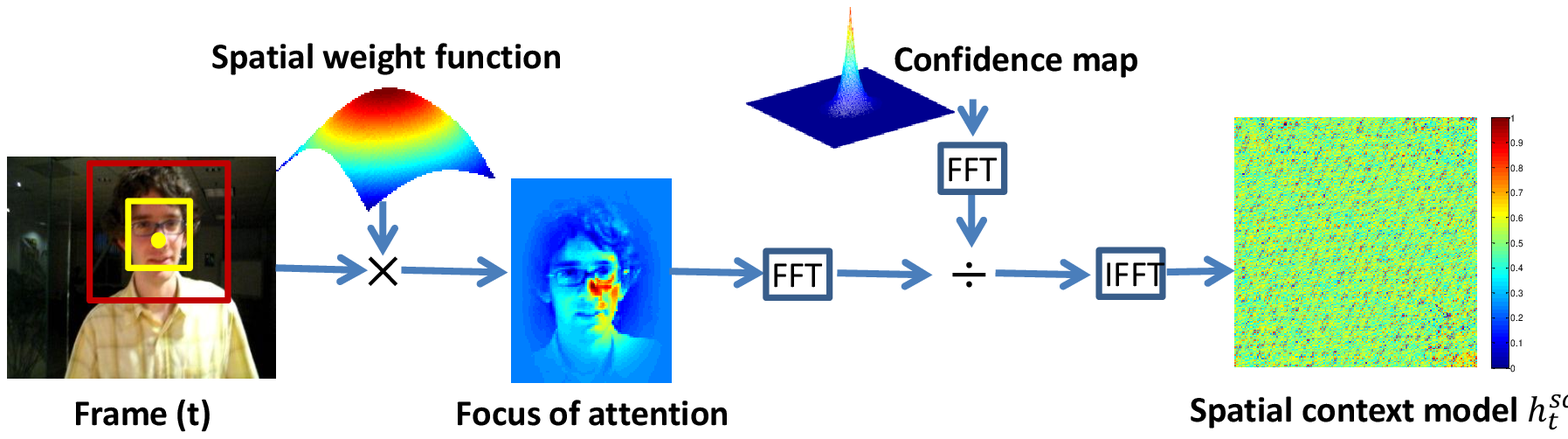}\\
\scriptsize
(a) Learn spatial context at the $t$-th frame\\
\includegraphics[width=1\linewidth]{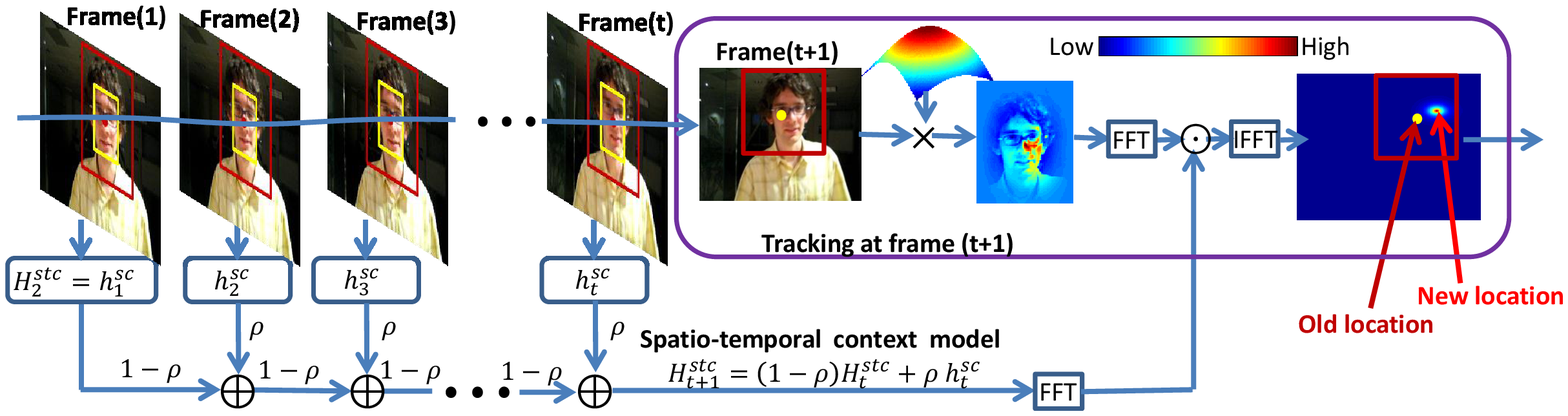}\\
\scriptsize
(b) Detect object location at the $(t\rm{+}1)$-th frame\\
\end{center}
\caption{Basic flow of our tracking algorithm. The local context
  regions are inside the red rectangles while the target locations are
  indicated by the yellow rectangles. FFT denotes the Fast Fourier Transform
  and IFFT is the inverse FFT.}
\label{fig:basicflow}
\end{figure*}

In visual tracking, a local context consists of a target object and
its immediate surrounding background within a determined region
(See the regions inside the red rectangles in Figure~\ref{fig:demoocc}).
Therefore, there exists a strong spatio-temporal relationship between
the local scenes containing the object in consecutive frames.
For instance, the target in
Figure~\ref{fig:demoocc} undergoes heavy occlusion which makes the
object appearance change significantly.
However, the local context containing the object does not change much
as the overall appearance remains similar and only a small part of the
context region is occluded.
Thus, the presence of local context in the current frame helps
predict the object location in the next frame.
This temporally proximal information in consecutive frames is
the temporal context which has been recently applied to object
detection~\cite{divvala2009empirical}.
%
Furthermore, the spatial relation between an object and its
local context provides specific information about the configuration
of a scene (See middle column in Figure~\ref{fig:demoocc}) which helps
discriminate the target from background when its appearance changes
much.
Recently, several methods~\cite{yang2009context, grabner2010tracking,
  dinh2011context, Wen_ECCV_2012}
exploit context information to facilitate visual tracking with
demonstrated success.
However, these approaches require high computational loads for feature
extraction in training and tracking phases.

In this paper, we propose a fast and robust tracking algorithm which
exploits spatio-temporal local context
information.
Figure~\ref{fig:basicflow} illustrates the basic flow of
our algorithm.
First, we learn a spatial context model between
the target object and its local surrounding background based on their
spatial correlations in a scene by solving a deconvolution
problem.
Next, the learned spatial context model is used to
update a spatio-temporal context model for the next frame.
Tracking in the next frame is formulated by computing a
confidence map as a convolution problem that integrates the
spatio-temporal context information, and the best object
location can be estimated by maximizing the confidence map (See
Figure~\ref{fig:basicflow} (b)).
Experiments on numerous challenging
sequences demonstrate that the proposed algorithm performs favorably
against state-of-the-art methods in terms of accuracy, efficiency and
robustness.

\section{Problem Formulation}
\begin{figure}[t]
\begin{center}
\includegraphics[width=0.45\linewidth]{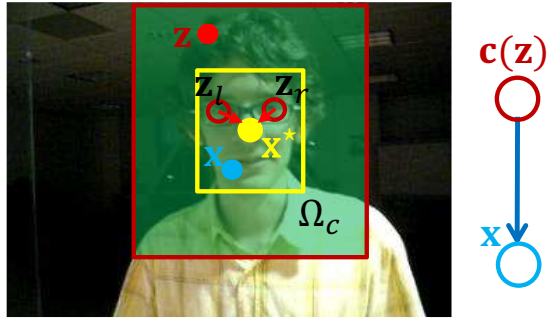}
\end{center}
\caption{Graphical model representation of spatial relationship
between object and its local context.
The local context region $\Omega_c$ is inside the red rectangle which
includes object region surrounding by the yellow rectangle
centering at the tracked result $\textbf{x}^\star$.
The context feature at location $\textbf{z}$ is denoted by
 $\textbf{c}(\textbf{z})=(I(\textbf{z}),\textbf{z})$ including
a low-level appearance representation (i.e., image intensity
$I(\textbf{z})$) and location information.}
\label{fig:graphrepresentation}
\end{figure}

The tracking problem is formulated by computing a confidence map which
 estimates the object location likelihood:
\begin{equation}
c(\textbf{x})=P(\textbf{x}|o),
 \label{eq:confmap}
\end{equation}
where $\textbf{x}\in\mathbb{R}^2$ is an object location and $o$ denotes
the object present in the scene.
In the following, the spatial context information is used to estimate
(\ref{eq:confmap}) and
Figure~\ref{fig:graphrepresentation} shows its graphical model
representation.

In the current frame, we have the object location $\textbf{x}^{\star}$
(i.e., coordinate of the tracked object center).
The context feature set is defined as
$X^c=\{\textbf{c}(\textbf{z})=(I(\textbf{z}),\textbf{z})|
\textbf{z}\in\Omega_c(\textbf{x}^\star)\}$
where $I(\textbf{z})$ denotes image intensity at location $\textbf{z}$
and $\Omega_c(\textbf{x}^\star)$ is the  neighborhood  of location
$\textbf{x}^\star$.
By marginalizing the joint probability
$P(\textbf{x},\textbf{c}(\textbf{z})|o)$, the object location likelihood
function in (\ref{eq:confmap}) can be computed by
\begin{equation}
\begin{aligned}
c(\textbf{x})&=P(\textbf{x}|o)\\
&=\textstyle\sum_{\textbf{c}(\textbf{z})\in X^c}P(\textbf{x},\textbf{c}(\textbf{z})|o)
\\
&=\textstyle\sum_{\textbf{c}(\textbf{z})\in
  X^c}P(\textbf{x}|\textbf{c}(\textbf{z}),o)P(\textbf{c}(\textbf{z})|o)
,
\end{aligned}
 \label{eq:marginalizingconfmap}
\end{equation}
where the conditional probability
$P(\textbf{x}|\textbf{c}(\textbf{z}),o)$ models the spatial
relationship between the object location and its context information which
helps resolve ambiguities when the image measurements allow
different interpretations, and
$P(\textbf{c}(\textbf{z})|o)$ is a context prior probability which
models appearance of the local context.
The main task in this work is to learn
$P(\textbf{x}|\textbf{c}(\textbf{z}),o)$ as it bridges the gap
between object location and its spatial context.
\subsection{Spatial Context Model}
The conditional probability function
$P(\textbf{x}|\textbf{c}(\textbf{z}),o)$ in
(\ref{eq:marginalizingconfmap}) is defined as
\begin{equation}
P(\textbf{x}|\textbf{c}(\textbf{z}),o)=h^{sc}(\textbf{x}-\textbf{z}),
 \label{eq:spatialcontext}
\end{equation}
where $h^{sc}(\textbf{x}-\textbf{z})$ is a function
(See Section~\ref{sec:fast-learning}) with respect to
the relative distance and direction between object
location $\textbf{x}$ and its local context location $\textbf{z}$,
thereby encoding the spatial relationship between an object and its
spatial context.

Note that  $h^{sc}(\textbf{x}-\textbf{z})$ is not a radially
symmetric function (i.e., $h^{sc}(\textbf{x}-\textbf{z})\neq
h^{sc}(|\textbf{x}-\textbf{z}|)$), and takes into account
different spatial relationships between an object and its local contexts,
thereby helping resolve ambiguities when similar objects appear in close
proximity.
For example, when a method tracks an eye based only on appearance
(denoted by $\textbf{z}_l$) in the $davidindoor$ sequence shown in
Figure~\ref{fig:graphrepresentation}, the tracker may be easily
distracted to the right one (denoted by $\textbf{z}_r$)
because both eyes and their surrounding backgrounds have
similar appearances (when the object moves fast and the search
region is large).
However, in the proposed method, while the locations of both eyes are
at similar distances to location $\textbf{x}^\star$
(here, it is location
of the context relative to object location $\textbf{z}_l$), their
relative locations to $\textbf{x}^\star$ are different,
thereby resulting in different spatial relationships,
i.e.,
$h^{sc}(\textbf{z}_{l}-\textbf{x}^\star)\neq
h^{sc}(\textbf{z}_{r}-\textbf{x}^\star)$.
That is, the non-radially symmetric function $h^{sc}$
helps resolve ambiguities effectively.

\subsection{Context Prior Model}
In (\ref{eq:marginalizingconfmap}), the context prior probability is
simply modeled by
\begin{equation}
P(\textbf{c}(\textbf{z})|o)=I(\textbf{z})w_{\sigma}(\textbf{z}-\textbf{x}^\star),
 \label{eq:context-prior}
\end{equation}
where $I(\cdot)$ is image intensity that represents appearance of
context and $w_{\sigma}(\cdot)$ is a weighted function defined by
\begin{equation}
w_{\sigma}(\textbf{z})=ae^{-\frac{|\textbf{z}|^2}{\sigma^2}},
 \label{eq:weight}
\end{equation}
%
where $a$ is a normalization constant that restricts
$P(\textbf{c}(\textbf{z})|o)$ in (\ref{eq:context-prior}) to range
from 0 to 1 that satisfies the definition of probability and $\sigma$
is a scale parameter.

In (\ref{eq:context-prior}), it models focus of attention that is
motivated by the biological visual system which
concentrates on certain image regions requiring detailed
analysis~\cite{torralba2003contextual}.
The closer the context
location $\textbf{z}$ is to the currently tracked target location
$\textbf{x}^\star$, the more important it is to predict the object
location in the coming frame, and larger weight should be
set.
Different from our algorithm that uses a spatially weighted function to
indicate the importance of context at different locations,
there exist other methods~\cite{belongie2002shape,wolf2006critical} in
which spatial sampling techniques are used to focus more detailed contexts
at the locations near the object center (i.e., the closer the location is
to the object center, the more context locations are sampled).

\subsection{Confidence Map}
\begin{figure}[t]
\begin{center}
\includegraphics[width=0.4\linewidth]{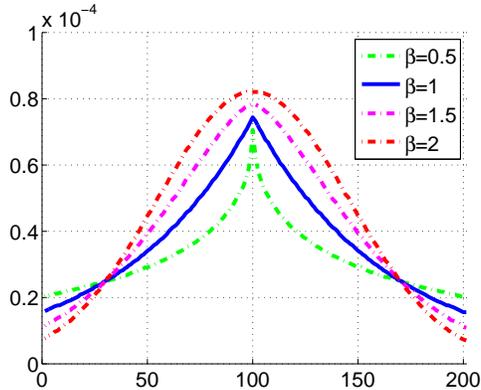}\\
\end{center}
   \caption{Illustration of 1-D cross section of the confidence map $c$(\textbf{x}) in (\ref{eq:confidencemap}) with different parameters $\beta$. Here, the object location $\textbf{x}^\star=(100,100)$.}
\label{fig:confidencemap}
\end{figure}

The confidence map of an object location is modeled as
\begin{equation}
c(\textbf{x})=P(\textbf{x}|o)=be^{-|\frac{\textbf{x}-\textbf{x}^\star}{\alpha}|^\beta},
 \label{eq:confidencemap}
\end{equation}
where $b$ is a normalization constant, $\alpha$ is a scale parameter
and $\beta$ is a shape parameter (See
Figure~\ref{fig:confidencemap}).

The object location ambiguity problem often occurs in visual tracking
which adversely affects tracking performance.
In~\cite{Babenko_PAMI_2011},
a multiple instance learning technique is adopted to handle the
location ambiguity problem with favorable tracking results.
The closer the location is to the currently tracked position, the
larger probability that the ambiguity occurs with (e.g., predicted
object locations that differ by a few pixels are all plausible
solutions and thereby cause ambiguities).
In our method, we resolve the location ambiguity problem by choosing a
proper shape parameter $\beta$.
As illustrated in Figure~\ref{fig:confidencemap}, a large $\beta$
(e.g., $\beta=2$) results in an oversmoothing effect for the
confidence value at locations near to the object center, thereby
failing to effectively reduce location ambiguities.
On the other hand, a small $\beta$ (e.g.,
$\beta=0.5$) yields a sharp peak near the object center,
thereby only activating much fewer positions when learning the spatial
context model.
This in turn may lead to overfitting in searching for the object
location in the coming frame.
We find that robust results can be obtained when $\beta=1$ in our
experiments.

\subsection{Fast Learning Spatial Context Model}
\label{sec:fast-learning}
Based on the confidence
map function (\ref{eq:confidencemap}) and the context prior model
(\ref{eq:context-prior}), our objective is to learn the spatial
context model (\ref{eq:spatialcontext}).
Putting (\ref{eq:confidencemap}),
(\ref{eq:context-prior}) and (\ref{eq:spatialcontext}) together,
we formulate (\ref{eq:marginalizingconfmap}) as
\begin{equation}
\begin{aligned}
c(\textbf{x})&=be^{-|\frac{\textbf{x}-\textbf{x}^\star}{\alpha}|^\beta}\\
&=\textstyle\sum_{\textbf{z}\in \Omega_c(\textbf{x}^\star)}
h^{sc}(\textbf{x}-\textbf{z})
I(\textbf{z})w_{\sigma}(\textbf{z}-\textbf{x}^\star)\\
&=h^{sc}(\textbf{x})\otimes(I(\textbf{x})
w_{\sigma}(\textbf{x}-\textbf{x}^\star)),
\end{aligned}
 \label{eq:convolutionconfmap}
\end{equation}
where $\otimes$ denotes the convolution operator.

We note (\ref{eq:convolutionconfmap}) can be transformed to the
frequency domain in which the Fast Fourier Transform (FFT)
algorithm~\cite{oppenheim1983signals} can be used for fast
convolution. That is,
\begin{equation}
\mathcal{F}(be^{-|\frac{\textbf{x}-\textbf{x}^\star}{\alpha}|^\beta})=
\mathcal{F}(h^{sc}(\textbf{x}))\odot
\mathcal{F}(I(\textbf{x})w_{\sigma}(\textbf{x}-\textbf{x}^\star)),
\label{eq:fft}
\end{equation}
where $\mathcal{F}$ denotes the FFT function and $\odot$ is
the element-wise product. Therefore, we have
\begin{equation}
h^{sc}(\textbf{x}) = \mathcal{F}^{-1}\bigg(
\frac{\mathcal{F}(be^{-|\frac{\textbf{x}-\textbf{x}^\star}{\alpha}|^\beta})}
{\mathcal{F}(I(\textbf{x})w_{\sigma}(\textbf{x}-\textbf{x}^\star))}
\bigg),
\label{eq:fftspatialcontextmodel}
\end{equation}
where $\mathcal{F}^{-1}$ denotes the inverse FFT function.

\section{Proposed Tracking Algorithm}
Figure~\ref{fig:basicflow} shows the basic flow of our
algorithm.
The tracking problem is formulated as a detection task.
We assume that the target location in the first frame has been
initialized manually or by some object detection algorithms.
At the $t$-th frame, we learn the spatial context model
$h_{t}^{sc}(\textbf{x})$ (\ref{eq:fftspatialcontextmodel}),
which is used to update the spatio-temporal context model
$H_{t+1}^{stc}(\textbf{x})$ (\ref{eq:spatiotemporal}) and applied
to detect the object location in the $(t\rm{+}1)$-th frame.
When the $(t\rm{+}1)$-th frame arrives, we crop out the local context region
$\Omega_{c}(\textbf{x}_{t}^\star)$ based on the tracked location
$\textbf{x}_{t}^\star$ at the $t$-th frame and construct the
corresponding context feature set
$X_{t+1}^c=\{\textbf{c}(\textbf{z})=
(I_{t+1}(\textbf{z}),\textbf{z})|\textbf{z}\in\Omega_c(\textbf{x}_t^\star)\}$.
The object location $\textbf{x}_{t+1}^\star$ in the
$(t\rm{+}1)$-th frame is determined by maximizing the new confidence map
\begin{equation}
\textbf{x}_{t+1}^\star=\mathop{\arg\max}_{\textbf{x}\in
 \Omega_c(\textbf{x}_t^\star)}c_{t+1}(\textbf{x}),
\label{eq:confmaptt}
\end{equation}
where $c_{t+1}(\textbf{x})$ is represented as
\begin{equation}
c_{t+1}(\textbf{x})=\mathcal{F}^{-1}\bigg(\mathcal{F}(H_{t+1}^{stc}(\textbf{x}))
\odot
\mathcal{F}(I_{t+1}(\textbf{x})w_{\sigma_t}(\textbf{x}-\textbf{x}_t^\star))\bigg),
\label{eq:confmapdeftt}
\end{equation}
which is deduced from (\ref{eq:fft}).
\subsection{Update of Spatio-Temporal Context}
\label{sec:stc}
The spatio-temporal context model is updated by
\begin{equation}
H_{t+1}^{stc}=(1-\rho)H_{t}^{stc}+\rho h_t^{sc},
\label{eq:spatiotemporal}
\end{equation}
where $\rho$ is a learning parameter and $h_t^{sc}$ is the spatial
context model computed by (\ref{eq:fftspatialcontextmodel}) at
the $t$-th frame.
We note (\ref{eq:spatiotemporal}) is a temporal filtering procedure
which can be easily observed in frequency domain
\begin{equation}
H_{\omega}^{stc} = F_{\omega}h_{\omega}^{sc},
\label{eq:temporalFrequency}
\end{equation}
where $H_{\omega}^{stc}\triangleq\int H_t^{stc}e^{-j\omega t}dt$ is
the temporal Fourier transform of $H_{t}^{stc}$ and similar to
$h_{\omega}^{sc}$. The temporal filter $F_{\omega}$ is formulated as
\begin{equation}
F_{\omega}=\frac{\rho}{e^{j\omega}-(1-\rho)},
\label{eq:filter}
\end{equation}
where $j$ denotes imaginary unit.
It is easy to validate that
$F_{\omega}$ in (\ref{eq:filter}) is a low-pass
filter~\cite{oppenheim1983signals}.
Therefore, our
spatio-temporal context model is able to effectively filter out
image noise introduced by appearance variations, thereby leading to
more stable results.
%
\subsection{Update of Scale}
According to (\ref{eq:confmapdeftt}), the target location in the
current frame is found by maximizing the confidence map derived from
the weighted context region surrounding the previous target
location.
However, the scale of the target often changes over
time.
Therefore, the scale parameter $\sigma$ in the weight function
$w_{\sigma}$ (\ref{eq:weight}) should be updated accordingly. We
propose the scale update scheme as
\begin{equation}
\label{eq:scaleupdate}
\left\{\begin{array}{rl}
s_{t}^{\prime}&= \sqrt{\frac{c_t(\textbf{x}_t^{\star})}
{c_{t-1}(\textbf{x}_{t-1}^{\star})}},\\
\overline{s}_t&=\frac{1}{n}\sum_{i=1}^{n}s_{t-i}^{\prime},\\
s_{t+1}&=(1-\lambda)s_t + \lambda\overline{s}_t,\\
\sigma_{t+1}&=s_t\sigma_t,
\end{array} \right.
\end{equation}
where $c_t(\cdot)$ is the confidence map that is computed by
(\ref{eq:confmapdeftt}), and $s_t^{\prime}$ is the estimated scale
between two consecutive frames.
To avoid oversensitive adaptation and
to reduce noise introduced by estimation error, the estimated target
scale $s_{t+1}$ is obtained through filtering in which
$\overline{s}_t$ is the average of the estimated scales from $n$
consecutive frames, and $\lambda>0$ is a fixed filter parameter
(similar to $\rho$ in (\ref{eq:spatiotemporal})).
The derivation details of~\eqref{eq:scaleupdate}
can be found in the Appendix~\ref{appen}.

\subsection{Analysis and Discussion}
\label{discussion}
We note that the low computational complexity is one prime
characteristic of the proposed algorithm in which only $6$ FFT
operations are involved for processing one frame including learning
the spatial context model (\ref{eq:fftspatialcontextmodel}) and
computing the confidence map (\ref{eq:confmapdeftt}).
The computational complexity for
computing each FFT is only $\mathcal{O}(MN\log(MN))$ for the local
context region of $M\times N$ pixels, thereby resulting in a fast
method ($350$ frames per second in MATLAB on an i$7$ machine).
More importantly, the proposed
algorithm achieves robust results as discussed bellow.
{\flushleft\textbf{Difference with related work.}}
It should be noted
that the proposed spatio-temporal context tracking algorithm is
significantly different from recently proposed context-based
methods~\cite{yang2009context,grabner2010tracking, dinh2011context,
  Wen_ECCV_2012} and approaches that use FFT for efficient
computation~\cite{bolme2009average, bolme2010visual,
  henriques2012circulant}.

All the above-mentioned context-based methods adopt some
strategies to find regions with consistent motion correlations to the
object.
In~\cite{yang2009context}, a data mining method is used to
extract segmented regions surrounding the object as auxiliary objects
for collaborative tracking.
To find consistent regions, key points surrounding the
object are first extracted to help locate the object
position~\cite{grabner2010tracking,dinh2011context,Wen_ECCV_2012}.
Next, SIFT or SURF descriptors are used to represent
these consistent regions~\cite{grabner2010tracking,
  dinh2011context,Wen_ECCV_2012}.
Thus, computationally expensive operations are required in
representing and finding consistent regions.
Moreover, due to the sparsity nature of key points, some
consistent regions that are useful for locating the object position
may be discarded.
In contrast, the proposed algorithm does not have these problems
because all the local regions surrounding the object are
considered as the potentially consistent regions, and the motion
correlations between the objects and its local contexts in consecutive
frames are learned by the spatio-temporal
context model that is efficiently computed by FFT.

In~\cite{bolme2009average,bolme2010visual}, the formulations are based
on correlation filters that are directly obtained by classic signal
processing algorithms.
At each frame, correlation filters are trained using a large
number of samples, and then combined to find the most correlated
position in the next frame.
In~\cite{henriques2012circulant}, the filters
proposed by~\cite{bolme2009average,bolme2010visual} are
kernelized and used to achieve more stable results.
The proposed algorithm is significantly different
from~\cite{bolme2009average, bolme2010visual, henriques2012circulant}
in several aspects.
First, our algorithm models the spatio-temporal relationships between
the object and its local contexts which is motivated by the human
visual system that uses context to help resolve ambiguities in complex
scenes efficiently and effectively.
Second, our algorithm focuses on the regions which require
detailed analysis, thereby effectively reducing the adverse effects of
background clutters and leading to more robust results.
Third, our algorithm handles the object location ambiguity problem
using the confidence map with a proper prior distribution,
thereby achieving more stable and accurate performance for visual
tracking.
Finally, our algorithm solves the scale adaptation problem but the
other FFT-based tracking methods~\cite{bolme2009average, bolme2010visual,
  henriques2012circulant} only track objects with a fixed scale
and achieve less accurate results than our method.
\begin{figure}[t]
\begin{center}
\includegraphics[width=0.5\linewidth]{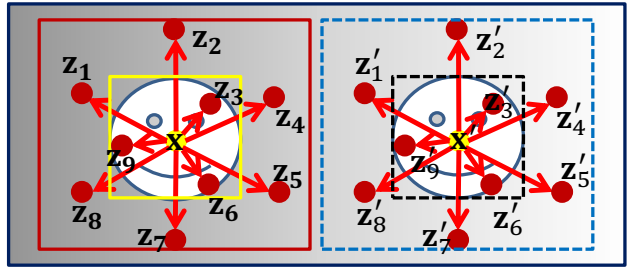}
\end{center}
   \caption{Illustration of why the proposed model is equipped to
handle distractor. The target inside the dotted rectangle is the distractor.
The different surrounding contexts (e.g., $\textbf{z}_i$ and $\textbf{z}_i^{\prime}$, $i=1,2,4,5,7,8$) can well discriminate target from distactor.}
\label{fig:distractor}
\end{figure}
%
{\flushleft\textbf{Robustness to occlusion and distractor.}}
\label{sec:distractor}
As shown in Figure~\ref{fig:demoocc}, the proposed algorithm handles
heavy occlusion well as most of context regions are not occluded
which have similar relative spatial relations (See middle column of
Figure~\ref{fig:demoocc}) to the target center, thereby helping
determine the target center.
%
Figure~\ref{fig:distractor} illustrates
our method is robust to distractor (i.e., the right object).
If tracking the target only based on its appearance information, the tracker will be distracted to the right one because of their similar appearances.
Although the distractor has similar appearance to the target, most of their
surrounding contexts have different appearances (See locations
$\textbf{z}_i,\textbf{z}_i^{\prime},i=1,2,4,5,7,8$) which are useful
to discriminate  target from distractor.
\section{Experiments}

We evaluate the proposed tracking algorithm based on spatio-temporal
context (STC) algorithm using $18$ video sequences with challenging factors
including heavy occlusion, drastic illumination changes, pose and
scale variation, non-rigid deformation, background cluster and motion
blur.
We compare the proposed STC tracker with $18$ state-of-the-art
methods.
The parameters of the proposed algorithm are {\textit{fixed}} for all
the experiments.
For other trackers, we use either the original source or binary codes
provided in which parameters of each tracker are tuned for best
results.
The $18$ trackers we compare with
are: scale mean-shift (SMS) tracker~\cite{collins2003mean}, fragment
tracker (Frag)~\cite{Adam_CVPR_2006}, semi-supervised
Boosting tracker (SSB)~\cite{Grabner_ECCV_2008}, local orderless
tracker (LOT)~\cite{oron2012locally}, incremental visual tracking
(IVT) method~\cite{Ross_IJCV_2008}, online AdaBoost tracker
(OAB)~\cite{Grabner_BMVC_2006}, multiple instance learning tracker
(MIL)~\cite{Babenko_PAMI_2011}, visual tracking decomposition method
(VTD)~\cite{Kwon_CVPR_2010}, L1 tracker (L1T)~\cite{Mei_PAMI_2011},
tracking-learning-detection (TLD) method~\cite{Kalal_CVPR_2010},
distribution field tracker (DF)~\cite{sevilla2012distribution},
multi-task tracker (MTT)~\cite{zhang2012robust}, structured output
tracker (Struck)~\cite{Hare_ICCV_2011}, context tracker
(ConT)~\cite{dinh2011context}, minimum output sum of square error
(MOS) tracker~\cite{bolme2010visual}, compressive tracker
(CT)~\cite{Zhang_ECCV_2012}, circulant structure tracker
(CST)~\cite{henriques2012circulant} and local-global tracker
(LGT)~\cite{cehovin2013robust}.
For the trackers involving randomness,
we repeat the experiments $10$ times on each sequence and
report the averaged results.
Implemented in MATLAB, our tracker runs
at $350$ frames per second (FPS) on an i$7$ $3.40$
GHz machine with $8$ GB RAM.
%
The MATLAB source codes will be released.
\subsection{Experimental Setup}
The size of context region is initially set to twice the size of the
target object.
The parameter $\sigma_t$ of~\eqref{eq:scaleupdate}
is initially set to $\sigma_1=\frac{s_h+s_w}{2}$, where $s_h$ and
$s_w$ are height and width
of the initial tracking rectangle, respectively.
The parameters of the map
function are set to $\alpha=2.25$ and $\beta=1$.
The learning parameter
$\rho=0.075$.
The scale parameter $s_t$ is initialized to $s_1=1$,
and the learning parameter $\lambda=0.25$.
The number of frames for updating the scale is set to $n=5$.
To reduce effects of illumination
change, each intensity value in the context region is
normalized by subtracting the average intensity of that region.
Then, the intensity in the context region multiplies a Hamming
window to reduce the frequency effect of image boundary
when using FFT~\cite{oppenheim1983signals,bolme2009average}.
%
%
\renewcommand{\tabcolsep}{2pt}
\begin{table}[t]\center
\caption{Success rate (SR)(\%). \textcolor{red}{\textbf{Red}} fonts
  indicate the best
  performance while the \textcolor{blue}{\textbf{blue}} fonts indicate
  the second best
  ones. The total number of evaluated frames is $7,591$.
}
\label{Table1}
{
\tiny
\begin{center}
\begin{tabular}{| c |c|c|c|c|c|c|c|c|c|c|c|c|c|c|c|c|c|c|c|}\hline
Sequence          &SMS~\cite{collins2003mean} &Frag~\cite{Adam_CVPR_2006} &SSB~\cite{Grabner_ECCV_2008} &LOT~\cite{oron2012locally} &IVT~\cite{Ross_IJCV_2008} &OAB~\cite{Grabner_BMVC_2006}     &MIL~\cite{Babenko_PAMI_2011}    &VTD~\cite{Kwon_CVPR_2010}    &L1T~\cite{Mei_PAMI_2011}    &TLD~\cite{Kalal_CVPR_2010}    &DF~\cite{sevilla2012distribution}     &MTT~\cite{zhang2012robust}  &Struck~\cite{Hare_ICCV_2011} &ConT~\cite{dinh2011context} &MOS~\cite{bolme2010visual} &CT~\cite{Zhang_ECCV_2012} &CST~\cite{henriques2012circulant} &LGT~\cite{cehovin2013robust} &\textbf{STC}    \\ \hline

{\textit{animal}}&13    &3	&51	&15	&4
&17	    &83 	&\textcolor{red}{\textbf{96}} 	&6	    &37	    &6	    &87	  &93	  &58	&3	   &92 &\textcolor{blue}{\textbf{94}} &7	 &\textcolor{blue}{\textbf{94}}\\

{\textit{bird}} &33      &64	&13	&5	&78	&\textcolor{red}{\textbf{94}}	&10	&9	&44	&42	 &\textcolor{red}{\textbf{94}}	 &10	 &48	&26	&11	 &8	 &47 &\textcolor{blue}{\textbf{89}}	&65\\

{\textit{bolt}}&58      &41	&18	&89	&15
&1	    &92	    &3	    &2	    &1	    &2	    &2	  &8	  &6	&25	   &\textcolor{blue}{\textbf{94}} &39	 &74 &\textcolor{red}{\textbf{98}}\\

{\textit{cliffbar}}&5 &24	&24	&26	&47
 &66	    &71  	&53	    &24	    &62	    &26	    &55	  &44	  &43	&6	   &\textcolor{blue}{\textbf{95}} &93 &81	 &\textcolor{red}{\textbf{98}}\\

{\textit{chasing}}&72   &77	&62	&20	&82
&71	    &65	    &70	    &72	    &76	    &70	    &95	  &85	  &53	&61	   &79 &\textcolor{blue}{\textbf{96}} &95	 &\textcolor{red}{\textbf{97}}\\

{\textit{car4}}&10   &34	&22	&1	&\textcolor{blue}{\textbf{97}}
 &30	    &37	    &35	    &94	    &88	    &26	    &22	  &96	  &90	&28	   &36 &44 &33	 &\textcolor{red}{\textbf{98}}\\

{\textit{car11}}&1   &1	&19	&32	&54
&14	    &48 	&25	    &46	    &67	    &78	    &59	  &18	  &47	 &\textcolor{blue}{\textbf{85}}	   &36 &48 &16	&\textcolor{red}{\textbf{86}}\\

{\textit{cokecan}}&1 &3	 &38	&4	&3
&53	    &18	    &7	    &16	    &17	    &13	    &85	  &\textcolor{red}{\textbf{94}}	  &20	&2	   &30 &86	 &18 &\textcolor{blue}{\textbf{87}}\\

{\textit{downhill}}&81   &89	 &53	&6	   &87	&82	&33	&\textcolor{red}{\textbf{98}}	&66	&13	&94	&54	 &87	 &71	 &28	&82	&72	 &73 &\textcolor{blue}{\textbf{99}}\\

{\textit{dollar}}&55  &41	&38	   &40	&21
 &16	    &46	    &39	    &39	    &39
 &\textcolor{red}{\textbf{100}}	&39
 &\textcolor{red}{\textbf{100}}	  &\textcolor{red}{\textbf{100}}
 &\textcolor{blue}{\textbf{89}}	   &87 &\textcolor{red}{\textbf{100}}
 &\textcolor{red}{\textbf{100}}	 &\textcolor{red}{\textbf{100}}\\

{\textit{davidindoor}}&6    &1	&36	 &20	&7
&24	    &30	    &38	    &18	    &\textcolor{blue}{\textbf{96}}	    &64	    &94	  &71	  &82	&43	   &46 &2 &95	 &\textcolor{red}{\textbf{100}}\\

{\textit{girl}}&7   &70	    &49	    &91 	&64
&68	    &28	    &68	    &56	    &79	    &59	    &71	  &\textcolor{blue}{\textbf{97}}	  &74	&3	   &27 &43 &51	 &\textcolor{red}{\textbf{98}}\\

{\textit{jumping}} &2 &34 	&81	    &22	    &\textcolor{red}{\textbf{100}}
 &82	    &\textcolor{red}{\textbf{100}}	&\textcolor{blue}{\textbf{87}}	    &13	    &76	    &12	    &\textcolor{red}{\textbf{100}}  &18	 &\textcolor{red}{\textbf{100}}	&6 	   &\textcolor{red}{\textbf{100}}	 &\textcolor{red}{\textbf{100}} &5	&\textcolor{red}{\textbf{100}}\\

{\textit{mountainbike}}&14 &13	&82	  &71	&\textcolor{red}{\textbf{100}}
&\textcolor{blue}{\textbf{99}}	&18
&\textcolor{red}{\textbf{100}}	&61	    &26	    &35
&\textcolor{red}{\textbf{100}}  &98	  &25	&55	   &89
&\textcolor{red}{\textbf{100}} &74
&\textcolor{red}{\textbf{100}}\\

{\textit{ski}}&22 &5	&65	&55	&16	&58	&33	&6	&5	&36	&6	&9	&\textcolor{red}{\textbf{76}}	 &43	&1	 &60	 &1 &\textcolor{blue}{\textbf{71}}	 &68\\

{\textit{shaking}}&2  &25	&30	&14	&1
&39	&83	    &\textcolor{red}{\textbf{98}}	    &3	    &15	    &84	    &2	  &48	  &12	&4	   &84 &36	 &48 &\textcolor{blue}{\textbf{96}}\\

{\textit{sylvester}}&70 &34	&67	&61	&45
&66	&77	    &33	    &40	    &\textcolor{red}{\textbf{89}}	    &33	    &68	  &81	  &84	&6	   &77 &84 &\textcolor{blue}{\textbf{85}}	&78\\

{\textit{woman}}&52      &27	&30	    &16	  &21
&18	&21	    &35	    &8	    &31	    &93	    &19	  &\textcolor{blue}{\textbf{96}}	  &28	&2	   &19 &21 &66	 &\textcolor{red}{\textbf{100}}\\\hline

Average SR  &35 &35     &45     &35     &49    &49    &52    &49    &40    &62    &53    &59    &\textcolor{blue}{\textbf{75}}    &62 &26    &62    &60 &68   &\textcolor{red}{\textbf{94}}\\\hline

\end{tabular}
\end{center}
}
\end{table}

\begin{table}[t]
\caption{Center location error (CLE)(in pixels) and
    average frame per second (FPS). \textcolor{red}{\textbf{Red}}
    fonts indicate the
    best performance while the \textcolor{blue}{\textbf{blue}} fonts
    indicate the second
    best ones. The total number of evaluated frames is $7,591$.
}
\label{Table2}
{
\tiny
\begin{center}
\center\begin{tabular}{| c |c|c|c|c|c|c|c|c|c|c|c|c|c|c|c|c|c|c|c|}\hline
Sequence          &SMS~\cite{collins2003mean} &Frag~\cite{Adam_CVPR_2006} &SSB~\cite{Grabner_ECCV_2008} &LOT~\cite{oron2012locally} &IVT~\cite{Ross_IJCV_2008} &OAB~\cite{Grabner_BMVC_2006}     &MIL~\cite{Babenko_PAMI_2011}    &VTD~\cite{Kwon_CVPR_2010}    &L1T~\cite{Mei_PAMI_2011}    &TLD~\cite{Kalal_CVPR_2010}    &DF~\cite{sevilla2012distribution}     &MTT~\cite{zhang2012robust}  &Struck~\cite{Hare_ICCV_2011} &ConT~\cite{dinh2011context} &MOS~\cite{bolme2010visual} &CT~\cite{Zhang_ECCV_2012} &CST~\cite{henriques2012circulant} &LGT~\cite{cehovin2013robust} &\textbf{STC}     \\ \hline

{\textit{animal}} &78   &100	  &25	&70	 &146
&62  &32     &17     &122    &125    &252    &17   &19      &76  &281   &18 &\textcolor{blue}{\textbf{16}}  &166 &\textcolor{red}{\textbf{15}}\\

{\textit{bird}}&25     &13	&101	&99	&13	 &\textcolor{red}{\textbf{9}}	&140	&57	&60	&145	 &12	 &156	 &21	&139	 &159	&79	&20	 &\textcolor{blue}{\textbf{11}} &15\\

{\textit{bolt}}&42     &43	&102	&\textcolor{blue}{\textbf{9}}	&65
 &227	    &\textcolor{blue}{\textbf{9}}	    &177	&261	&286	&277	&293  &149	   &126	&223   &10 &210  &12 &\textcolor{red}{\textbf{8}} \\

{\textit{cliffbar}}&41  &34	&56	&36	&37
&33	    &13	    &30	    &40	    &70	    &52	    &25	  &46	   &49	&104   &\textcolor{blue}{\textbf{6}}  &\textcolor{blue}{\textbf{6}}	 &10 &\textcolor{red}{\textbf{5}}\\

{\textit{chasing}}&13  &9	&44	&32	&6
 &9	    &13	    &23	    &9	    &47	    &31	    &\textcolor{blue}{\textbf{5}}	  &6	   &16	&68    &10 &\textcolor{blue}{\textbf{5}} &6	 &\textcolor{red}{\textbf{4}}\\

\textit{car4} &144    &56	&104	&177	&14
&109	    &63	    &127	&16	    &13	    &92	    &158  &\textcolor{red}{\textbf{9}}	   &\textcolor{blue}{\textbf{11}}	&117   &63 &44	 &47 &\textcolor{blue}{\textbf{11}}\\

{\textit{car11}}&86  &117	&11	&30	&\textcolor{blue}{\textbf{7}}
 &11	    &8	    &20	    &8	    &12	    &\textcolor{red}{\textbf{6}}	    &8	  &9	   &8 	&8	   &9  &8 &16	 &\textcolor{blue}{\textbf{7}}\\

{\textit{cokecan}}&60   &70	&15	&46	&64
&11	    &18	    &68	    &40	    &29	    &30	    &10	  &\textcolor{blue}{\textbf{7}}	   &36	&53	   &16 &9	 &32 &\textcolor{red}{\textbf{6}}\\

{\textit{downhill}}&14  &11	&102	&226	&22	    &12	  &117	&\textcolor{blue}{\textbf{9}}	 &35	 &255	 &10	   &77	 &10	   &62	&116	 &12	 &129	&12 &\textcolor{red}{\textbf{8}}\\

{\textit{dollar}}&55  &56	&66	&66	&23
&28	    &23	    &65	    &65	    &72	    &\textcolor{blue}{\textbf{3}}	    &71	  &18	   &5	&12	   &20 &5	 &4 &\textcolor{red}{\textbf{2}}\\

{\textit{davidindoor}}&176 &103	&45	&100	&281
&43	    &33	    &40	    &86	    &13	    &27	    &\textcolor{blue}{\textbf{11}}	  &20	   &22	&78	   &28 &149 &12	 &\textcolor{red}{\textbf{8}}\\

{\textit{girl}}&130   &26	&50	&12	&36
&22	    &34	    &41	    &51	    &23	    &27	    &23	  &\textcolor{red}{\textbf{8}}	   &34	&126   &39 &43	 &35 &\textcolor{blue}{\textbf{9}}\\

{\textit{jumping}}&63 &30	&11	&43	&\textcolor{blue}{\textbf{4}}
&11	    &\textcolor{blue}{\textbf{4}}	    &17	    &45	    &13	    &73	    &7	  &42	  &\textcolor{blue}{\textbf{4}}	 &155	&6	 &\textcolor{red}{\textbf{3}}	&89  &\textcolor{blue}{\textbf{4}}\\

{\textit{mountainbike}}&135   &209	&11	    &24	    &\textcolor{red}{\textbf{5}}
&11	&208	&7	    &74	    &213	&155	&7	  &8	   &149	&16	   &11 &\textcolor{red}{\textbf{5}} &12	 &\textcolor{blue}{\textbf{6}}\\

{\textit{ski}}&91  &134	&\textcolor{blue}{\textbf{10}}	&12	&51	&11	&15	&179	&161	&222	&147	&33	 &\textcolor{red}{\textbf{8}}	&78	&386	 &11	&237	&13 &12\\

{\textit{shaking}}&224    &55	    &133	&90	    &134
&22	&11	    &\textcolor{red}{\textbf{5}}	    &72	    &232	&11	    &115  &23	   &191	&194   &11 &21	 &33 &\textcolor{blue}{\textbf{10}}\\

{\textit{sylvester}}&15   &47	&14	    &23	    &138
&12	&9	    &66	    &49	    &\textcolor{blue}{\textbf{8}}	    &56	    &18	  &9	   &13	&65	   &9  &\textcolor{red}{\textbf{7}}	&11 &11\\

{\textit{woman}}&49  &118	    &86	    &131	&112
&120	&119	&110	&148	&108	&12	    &169  &\textcolor{red}{\textbf{4}}	   &55	&176   &122&160	 &23 &\textcolor{blue}{\textbf{5}}\\\hline
Average CLE    &79 &63     &54     &70     &84     &43    &43    &58    &62    &78    &52    &80    &\textcolor{blue}{\textbf{19}}    &42 &103    &29    &54  &22  &\textcolor{red}{\textbf{8}} \\\hline
Average FPS   &12 &7 &11 &0.7
&33 &22 &38 &5 &1 &28 &13 &1 &20 &15  &\textcolor{blue}{\textbf{200}} &90 &120 &8    &\textcolor{red}{\textbf{350}} \\ \hline
\end{tabular}
\end{center}
}
\end{table}
\subsection{Experimental Results}
\begin{figure*}[!ht]
\hspace{-.3cm}
\begin{center}
\begin{tabular}{cc}
\includegraphics[width=.16\linewidth]{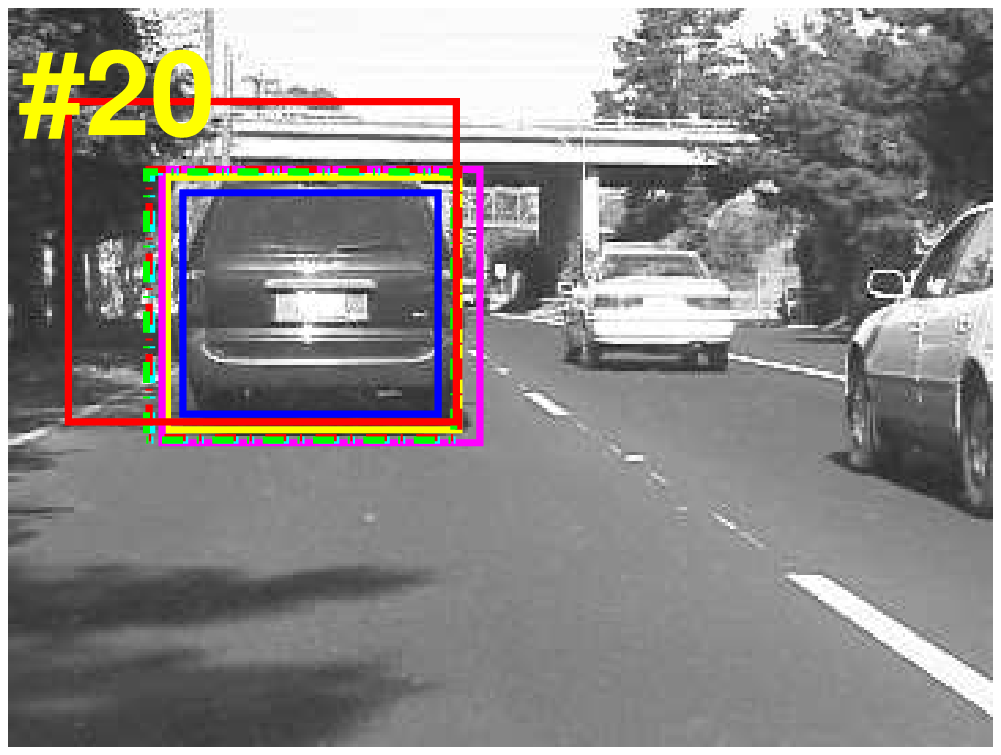}
\includegraphics[width=.16\linewidth]{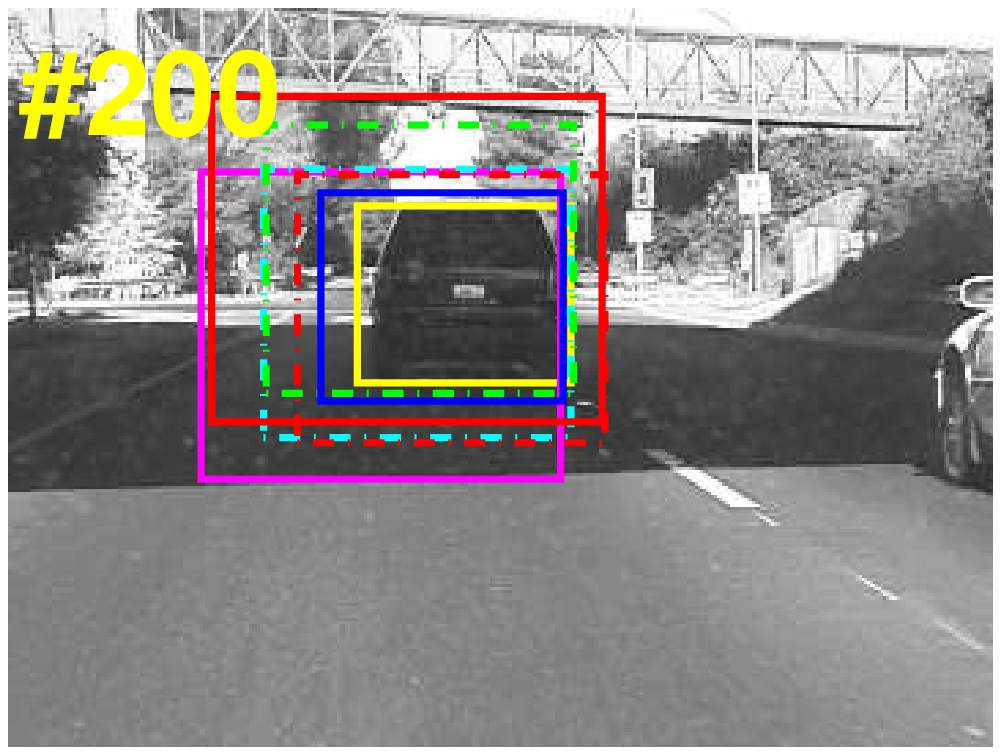}
\includegraphics[width=.16\linewidth]{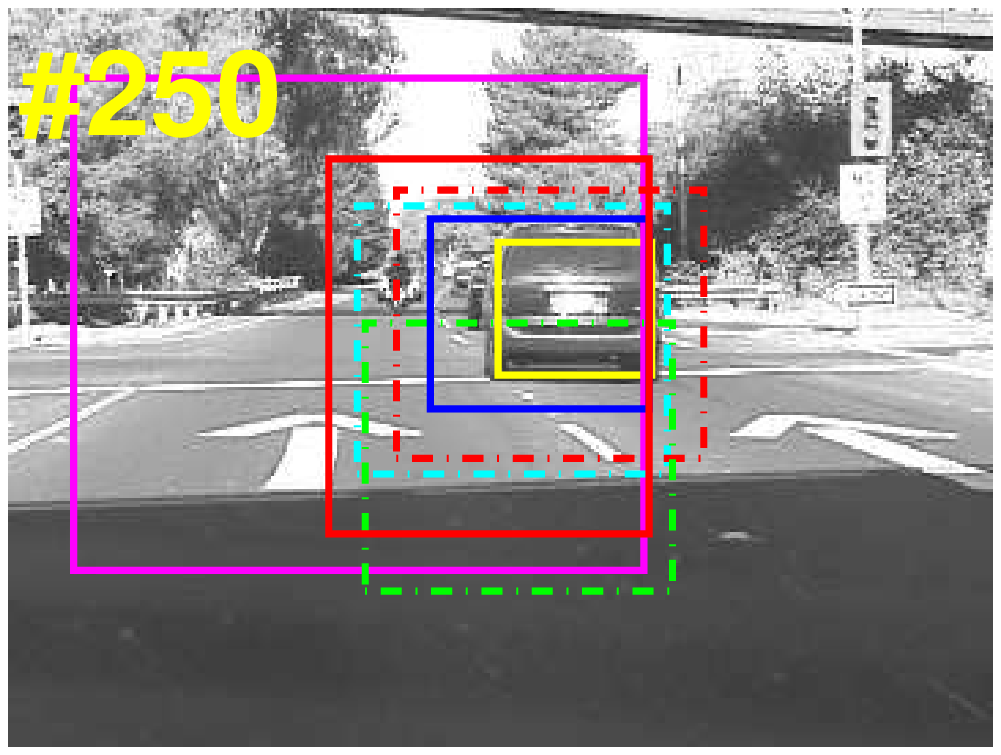}&
\includegraphics[width=.16\linewidth]{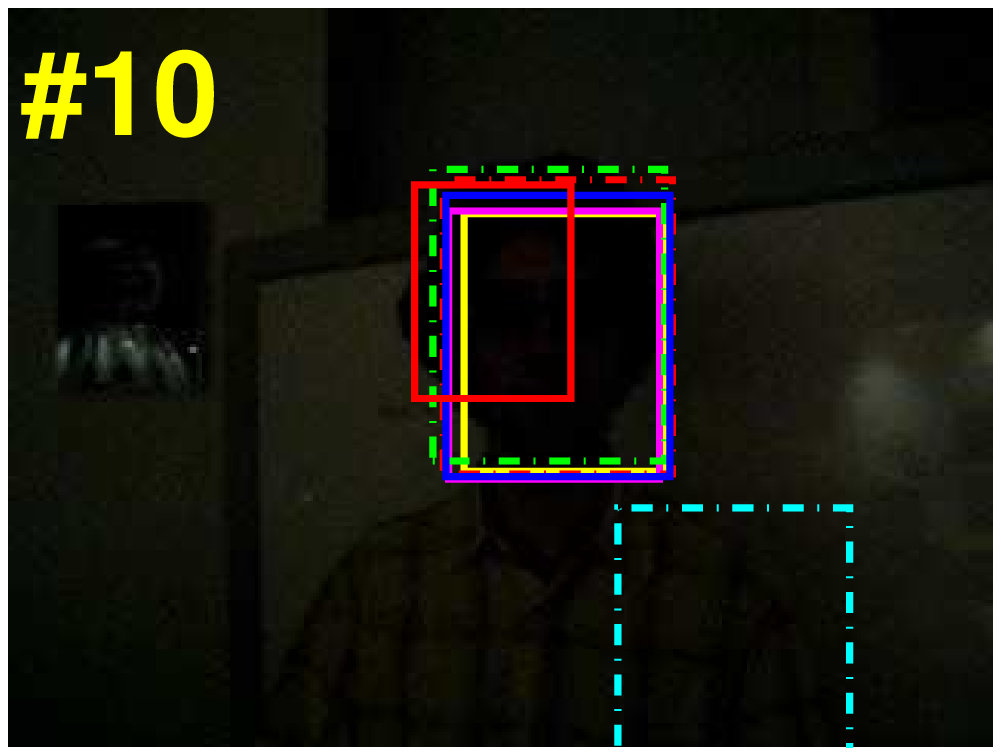}
\includegraphics[width=.16\linewidth]{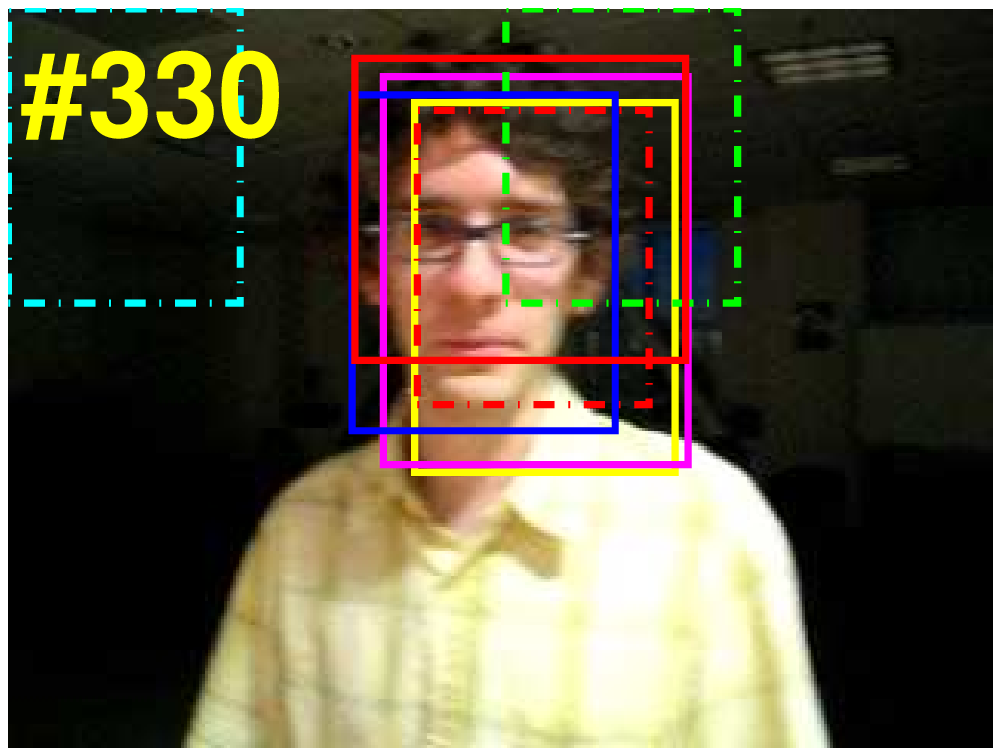}
\includegraphics[width=.16\linewidth]{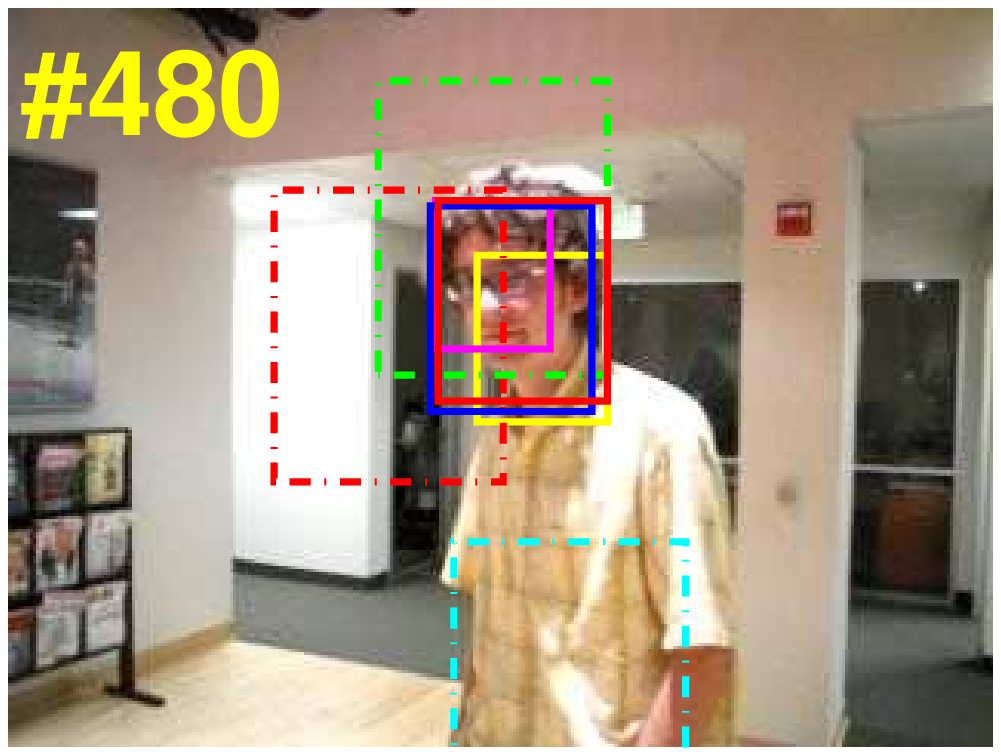}\\
\scriptsize (a) car4 & \scriptsize (b) davidindoor \\
\includegraphics[width=.16\linewidth]{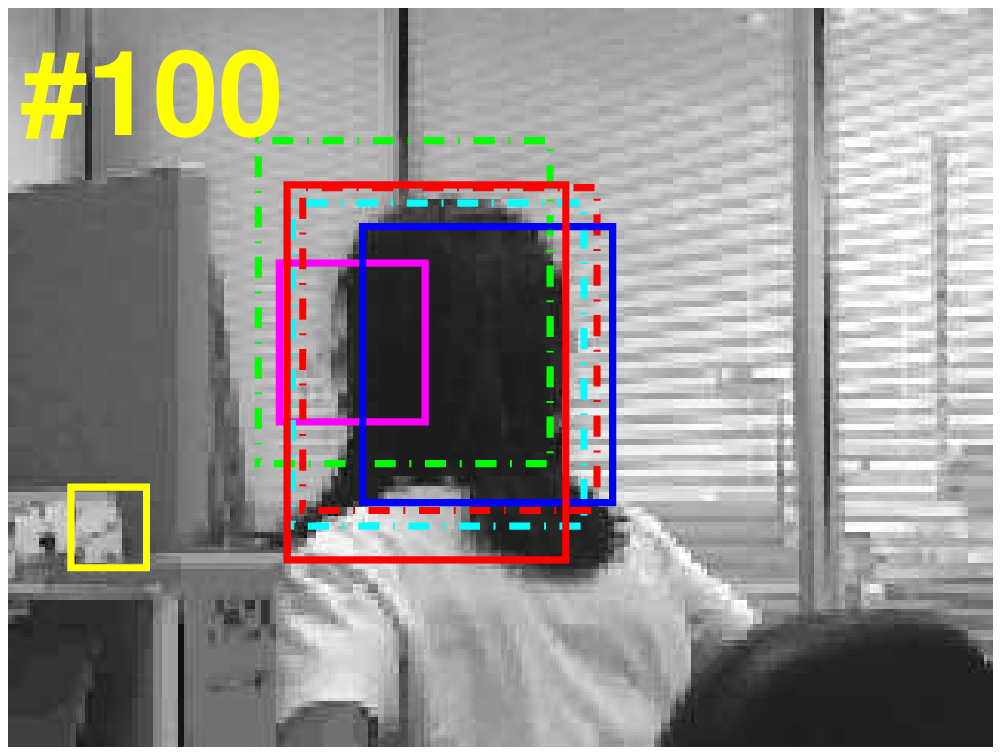}
\includegraphics[width=.16\linewidth]{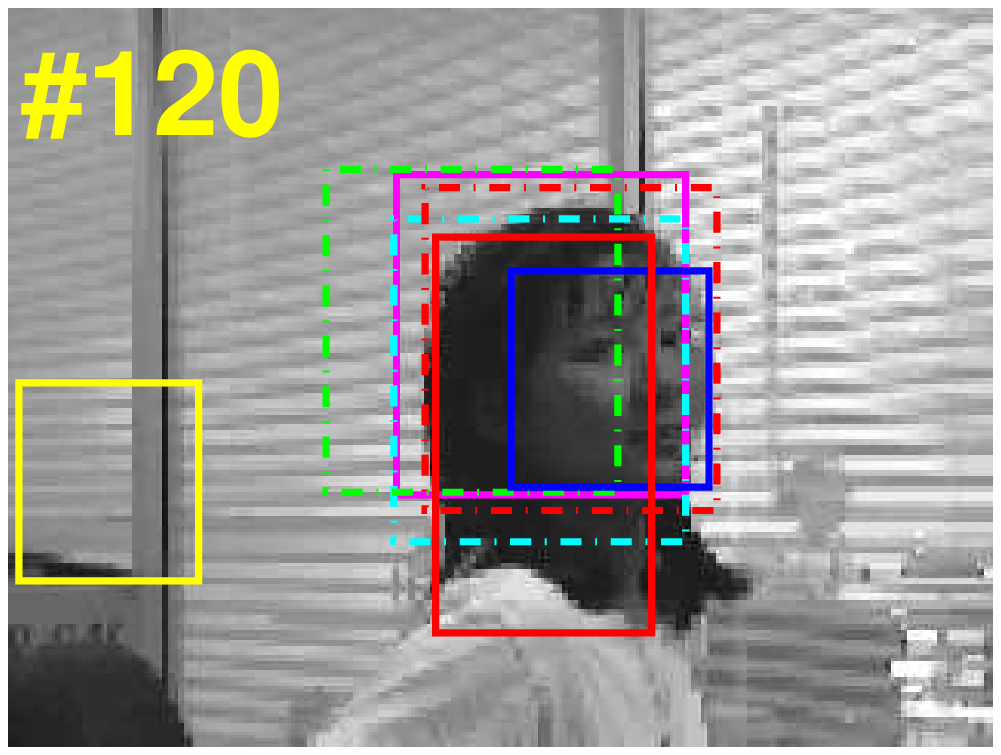}
\includegraphics[width=.16\linewidth]{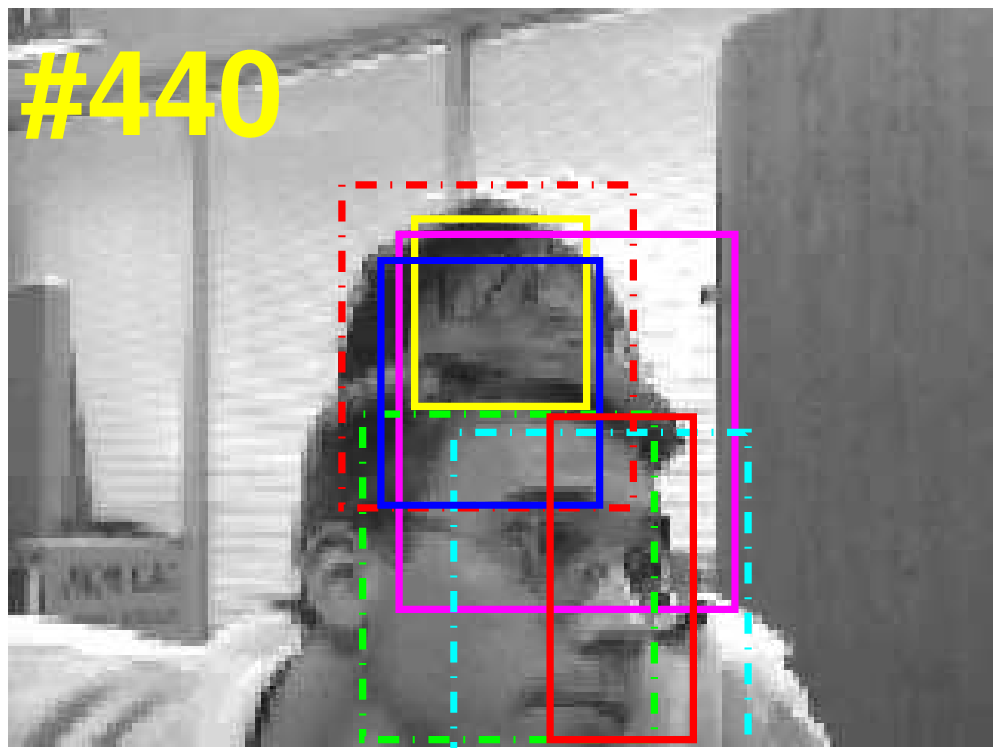}&
\includegraphics[width=.16\linewidth]{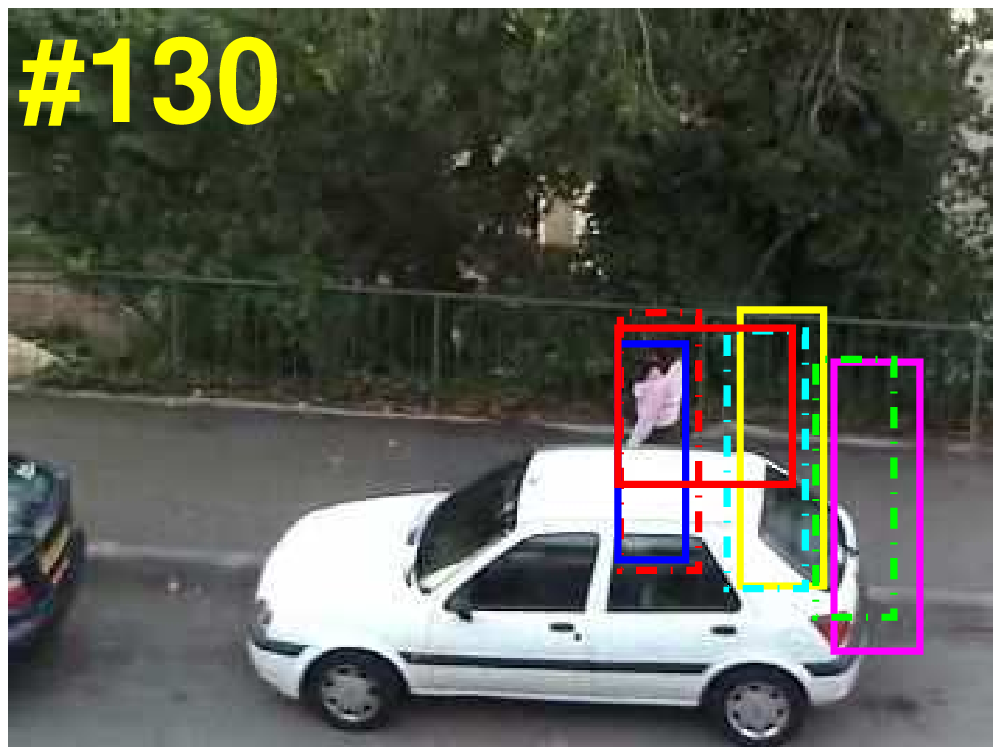}
\includegraphics[width=.16\linewidth]{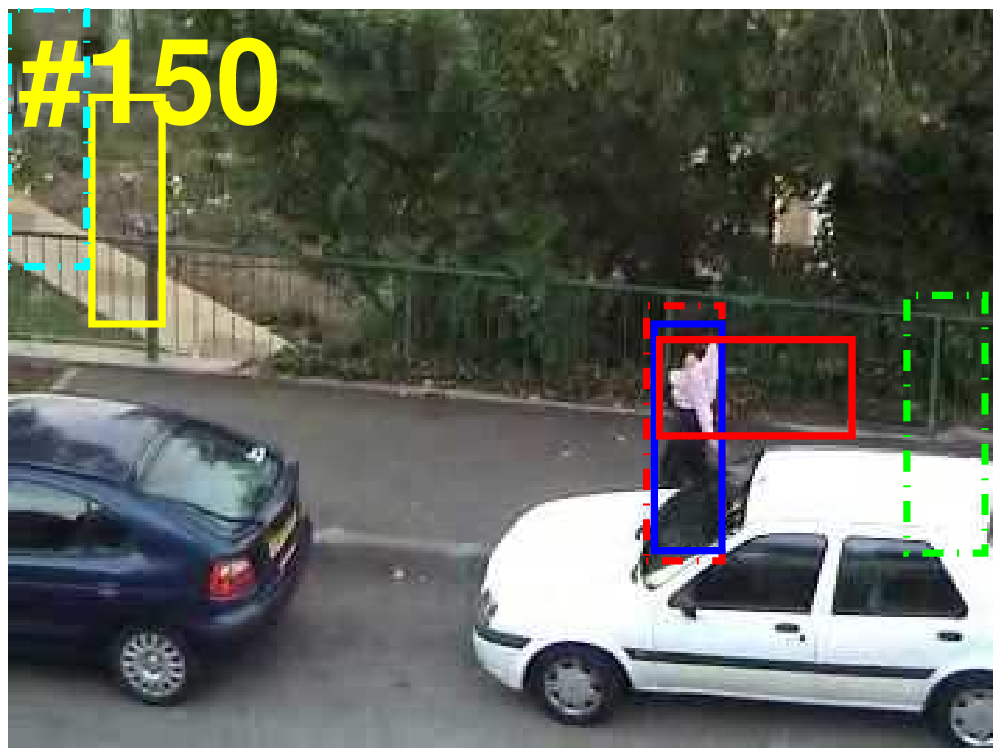}
\includegraphics[width=.16\linewidth]{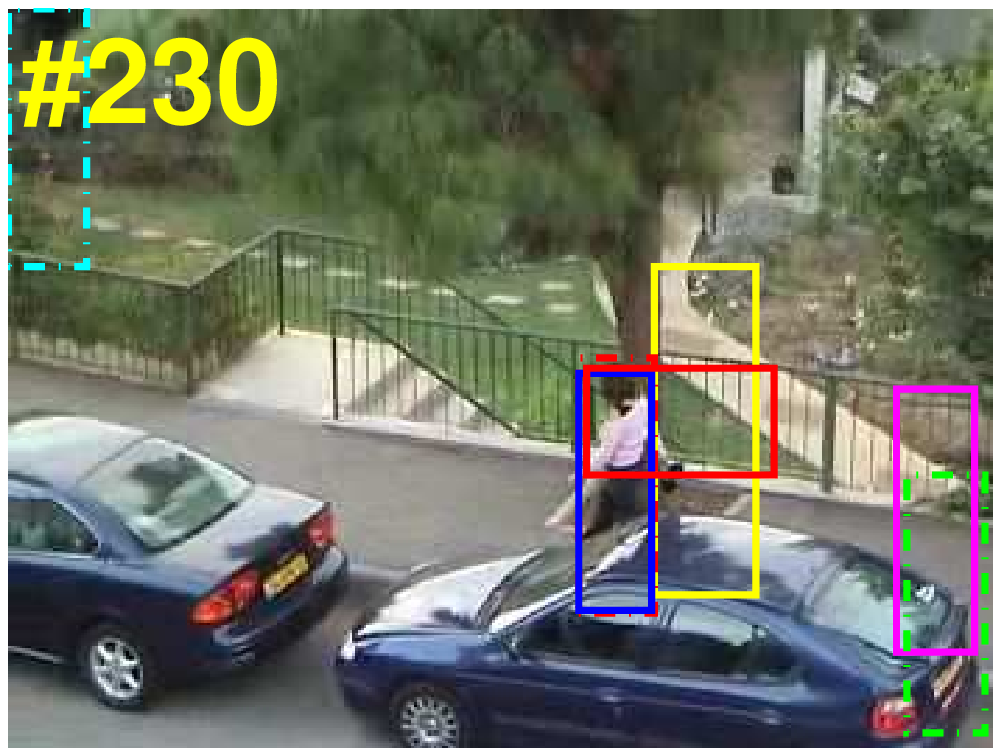}\\
\scriptsize (c) girl & \scriptsize (d) woman \\
\includegraphics[width=.16\linewidth]{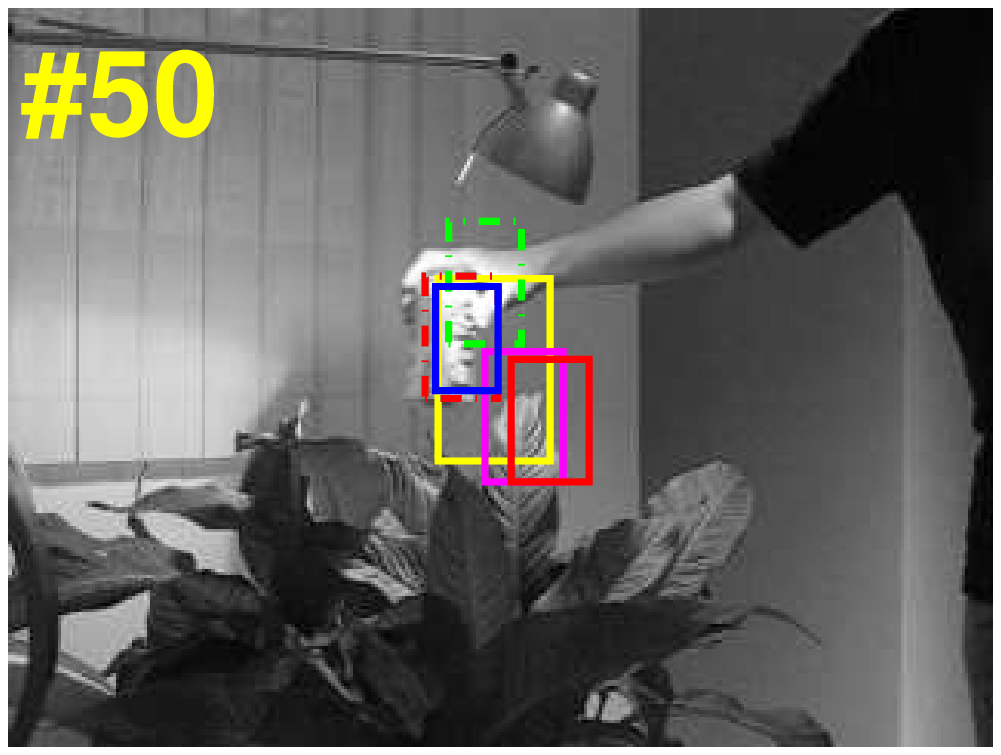}
\includegraphics[width=.16\linewidth]{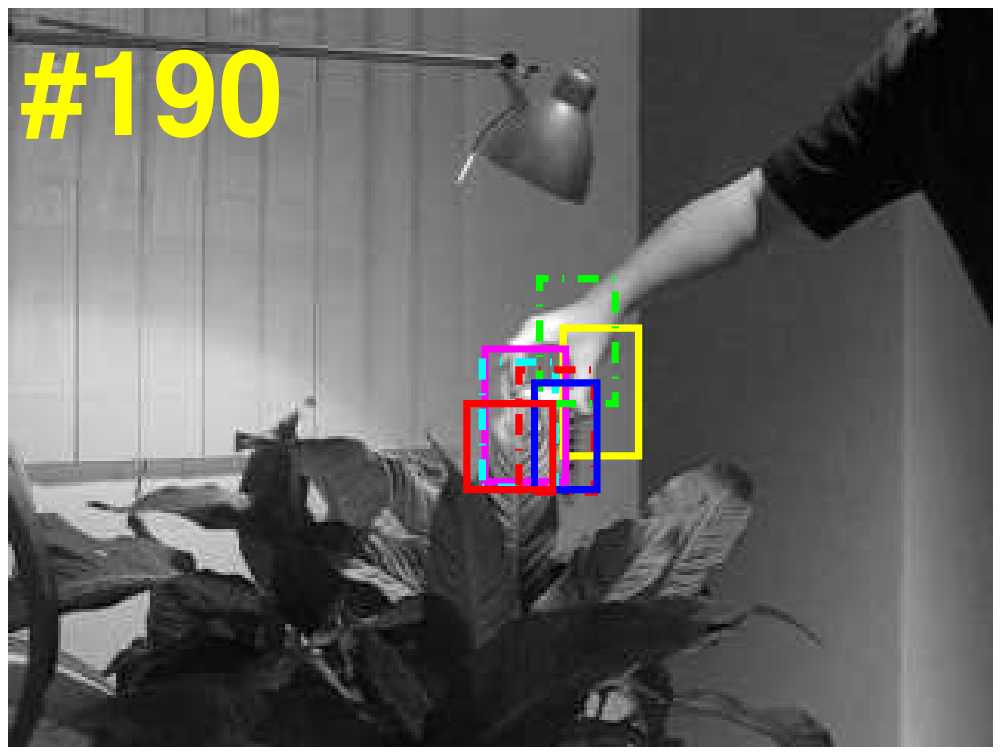}
\includegraphics[width=.16\linewidth]{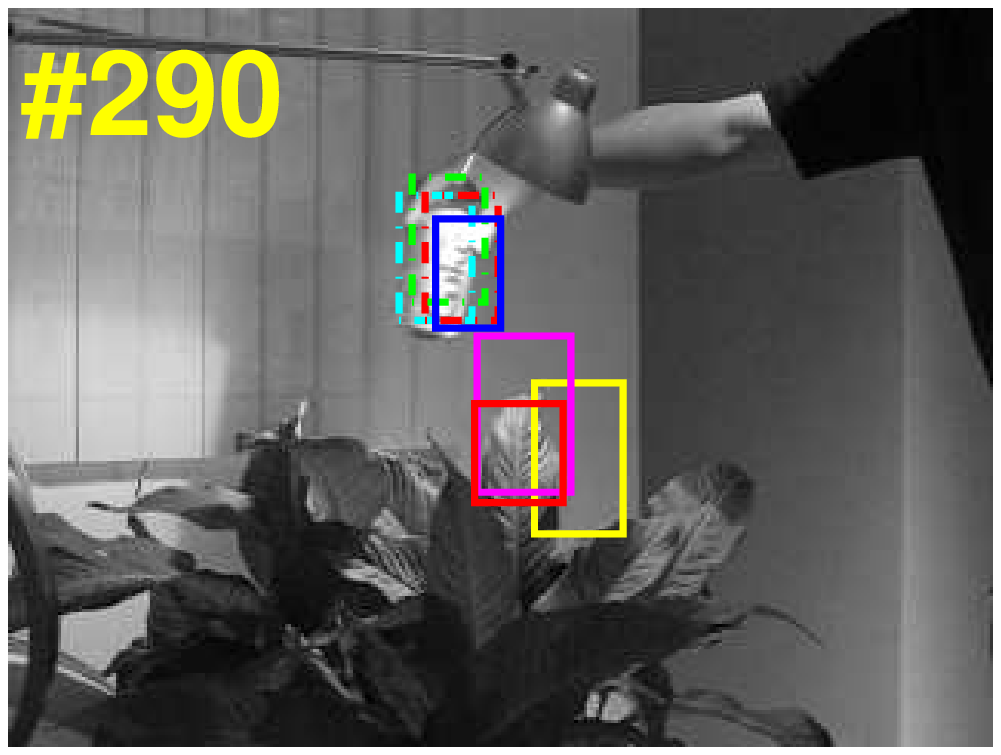}&
\includegraphics[width=.16\linewidth]{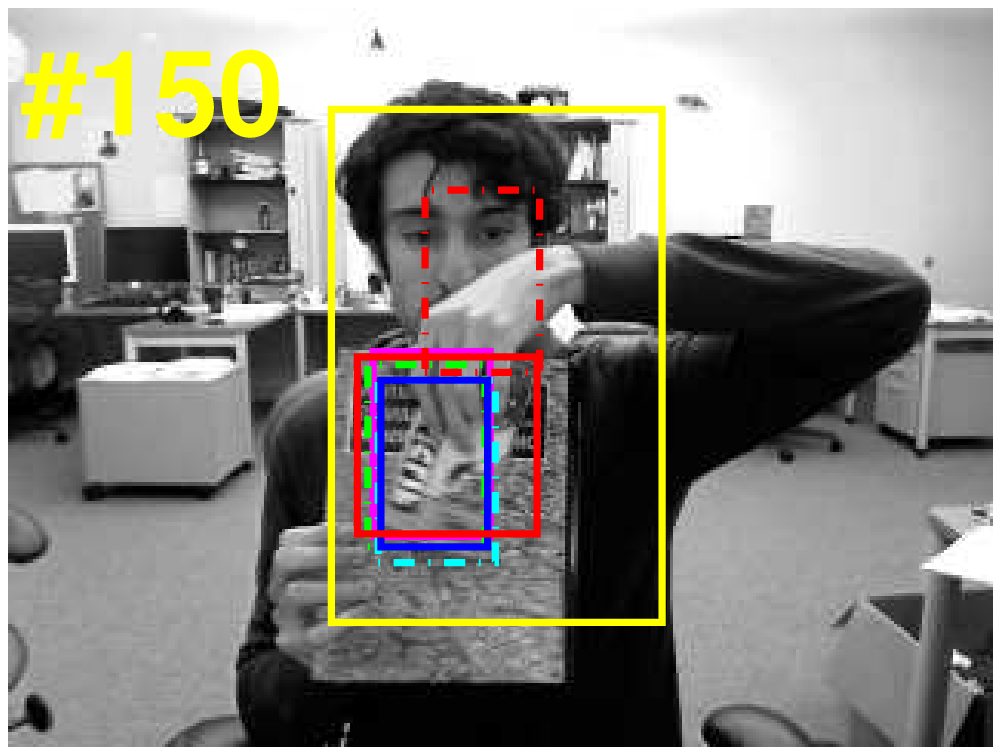}
\includegraphics[width=.16\linewidth]{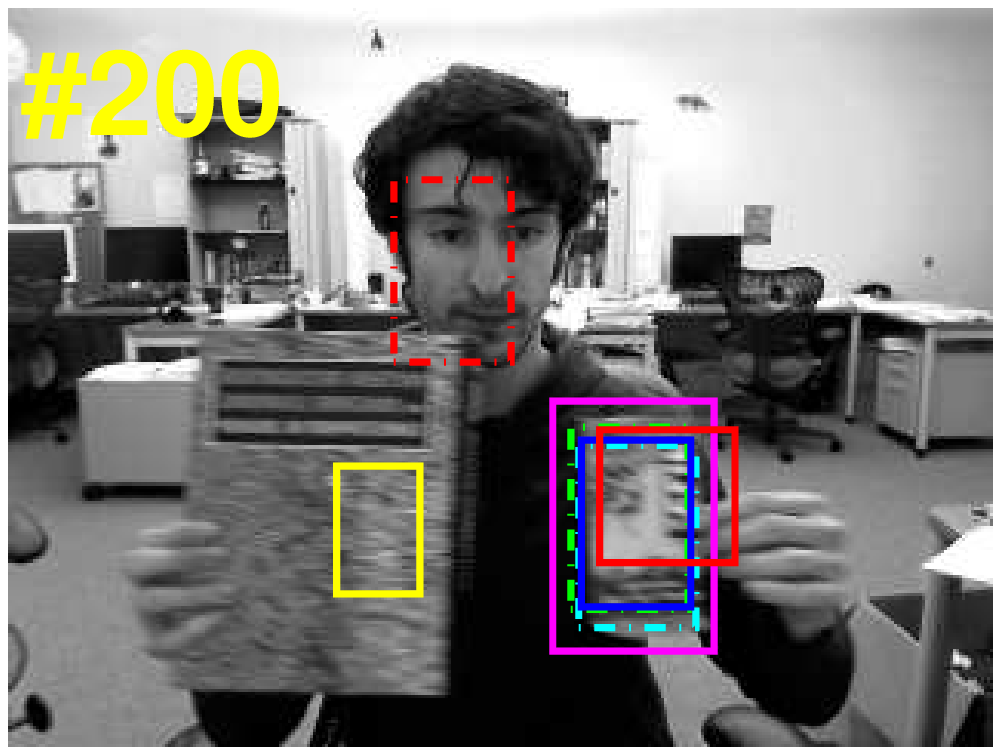}
\includegraphics[width=.16\linewidth]{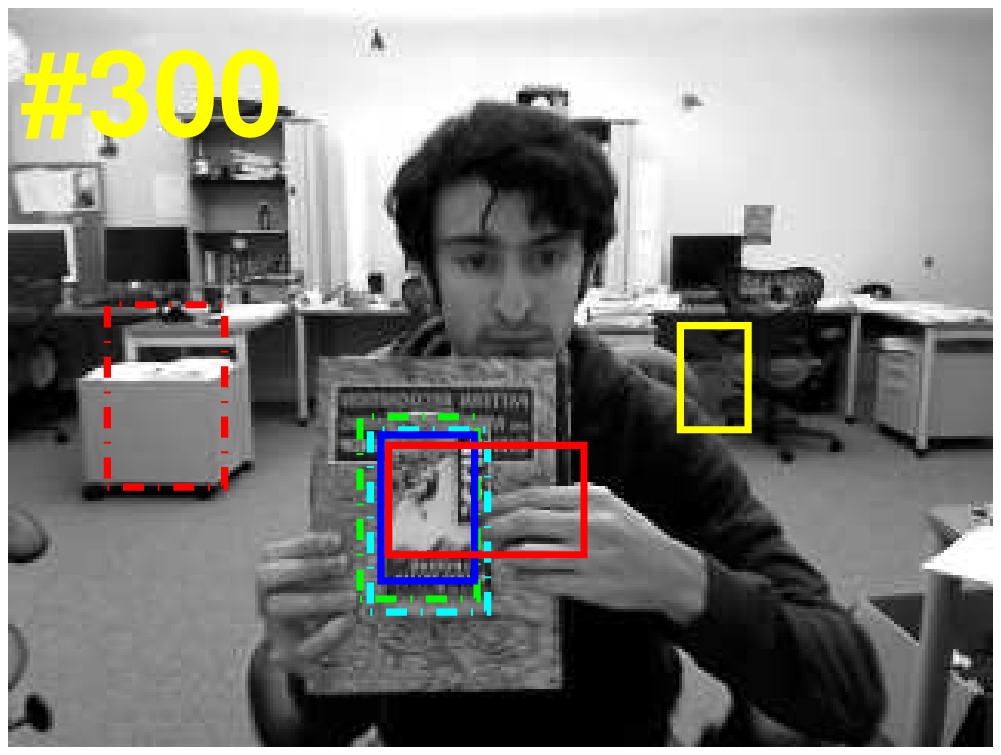}\\
\scriptsize (e) cokecan & \scriptsize (f) cliffbar \\
\end{tabular}
\includegraphics[width=.43\linewidth]{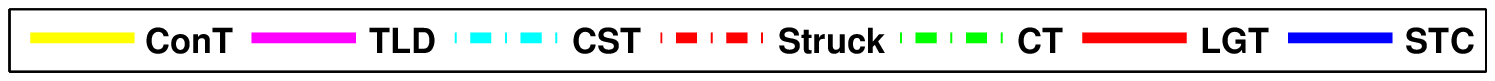}
\end{center}
\vspace{-2mm}
\caption{Screenshots of tracking results.}
\label{fig:screenshots}
\end{figure*}

We use two evaluation criteria to quantitatively evaluate the $19$
trackers: the center location error (CLE) and success rate (SR), both
computed based on the manually labeled ground truth results of each frame.
The score of success rate is defined as
$score=\frac{area(R_t\bigcap R_g)}{area(R_t\bigcup R_g)}$,
where $R_t$ is a tracked bounding box and $R_g$ is the ground
truth bounding box, and the result of one frame
is considered as a success if $score>0.5$.
Table~\ref{Table1} and Table~\ref{Table2}
show the quantitative results in which the proposed STC tracker
achieves the best or second best performance in most sequences both in
terms of center location error and success rate.
Furthermore, the proposed tracker is the most efficient ($350$ FPS on
average) algorithm among all evaluated methods.
Although the CST~\cite{henriques2012circulant} and
MOS~\cite{bolme2010visual} methods also use FFT for fast computation,
the CST method performs time-consuming kernel operations
and the MOS tracker computes several correlation filters in
each frame, thereby making these two approaches less efficient than the
proposed algorithm.
Furthermore, both CST and MOS methods only track target with fixed
scale, which achieve less accurate results that the proposed method
with scale adaptation.
Figure~\ref{fig:screenshots} shows some tracking results of different
trackers.
For presentation clarity, we only show the results of the top $7$
trackers in terms of average success rates.
%
{\flushleft\textbf{Illumination, scale and pose variation.}}
There are large illumination variations in the evaluated sequences.
The appearance of the target object in the~{\textit{car4}} sequence
changes significantly due to the cast shadows and ambient lights (See
$\#200, \#250$ in the~{\textit{car4}} sequence shown in
Figure~\ref{fig:screenshots}).
Only the models of the IVT, L1T, Struck and STC methods adapt to these
illumination variations well.
Likewise, only the VTD and our STC methods perform favorably
on the~{\textit{shaking}} sequence because the object appearance
changes drastically due to the stage lights and sudden pose variations.
The~{\textit{davidindoor}} sequence contain gradual pose and scale
variations as well as
illumination changes.
Note that most reported results using this sequence are only on
subsets of the available
frames, i.e., not from the very beginning of the~{\textit{davidindoor}}
video when the target face is in nearly complete darkness.
In this work, the full sequence is used to better evaluate the
performance of all algorithms.
Only the proposed algorithm is able to achieve favorable
tracking results on this sequence both in terms of accuracy and
success rate.
This can be attributed to the use of spatio-temporal
context information which facilitates filtering out noisy observations
(as discussed in Section~\ref{sec:stc}),
thereby enabling the proposed STC algorithm to relocate
the target when object appearance changes drastically due to
illumination, scale and pose variations.
{\flushleft\textbf{Occlusion, rotation, and pose variation.}}
The target
objects in the~{\textit{woman}}, {\textit{girl}} and {\textit{bird}} sequences are partially occluded at times.
The object in the {\textit{girl}} sequence
also undergoes in-plane rotation (See $\#100, \#120$ of
the {\textit{girl}} sequence in Figure~\ref{fig:screenshots}) which makes
the tracking tasks difficult.
Only the proposed algorithm
is able to track the objects successfully in most frames of this
sequence.
The~{\textit{woman}} sequence has non-rigid deformation and
heavy occlusion (See $\#130,\#150, \#230$ of the~{\textit{woman}} sequence
in Figure~\ref{fig:screenshots}) at the same time.
All the other
trackers fail to successfully track the object except the Struck
and the proposed STC algorithms.
As most of the local contexts surrounding the target objects are not
occluded in these sequences, such information facilitates the proposed algorithm
relocating the object even they are almost fully occluded (as discussed in Figure~\ref{fig:demoocc}).
{\flushleft\textbf{Background clutter and abrupt motion.}}
In the~{\textit{animal}}, {\textit{cokecan}} and {\textit{cliffbar}}
sequences, the target objects undergo fast movements in the cluttered
backgrounds.
The target object in the~{\textit{chasing}} sequence undergoes abrupt motion
with 360 degree out-of-plane rotation, and the proposed algorithm
achieves the best performance both in terms of success rate and center
location error.
The~{\textit{cokecan}} video contains a specular object with
in-plane rotation and heavy occlusion, which makes this tracking task
difficult.
Only the Struck and the proposed STC methods are able to
successfully track most of the frames.
In the {\textit{cliffbar}} sequence, the texture in the background is
very similar to that of the target object.
Most trackers drift to background except the CT, CST, LGT and
our methods (See $\#300$ of the {\textit{cliffbar}} sequence in
Figure~\ref{fig:screenshots}).
Although the target and its local background have very similar
texture, their spatial relationships and appearances of local contexts are different which are used by
the proposed algorithm when learning a confidence map (as discussed in Section~\ref{sec:distractor}).
Hence, the proposed STC algorithm is able to separate the target object
from the background based on the spatio-temporal context.

\section{Conclusion}
In this paper, we present a simple yet fast and robust
algorithm which exploits spatio-temporal context information for
visual tracking.
Two local context models (i.e., spatial context
and spatio-temporal context models) are proposed which are
robust to appearance variations introduced by occlusion, illumination
changes, and pose variations.
The Fast Fourier Transform algorithm is
used in both online learning and detection, thereby resulting in
an efficient tracking method that runs at $350$ frames per
second with MATLAB implementation.
Numerous experiments with state-of-the-art algorithms on
challenging sequences demonstrate that the proposed algorithm achieves
favorable results in terms of accuracy, robustness, and speed.
\section*{\bf Appendex}
    \label{appen}

\begin{figure}[t]
\begin{center}
\includegraphics[width=0.7\linewidth]{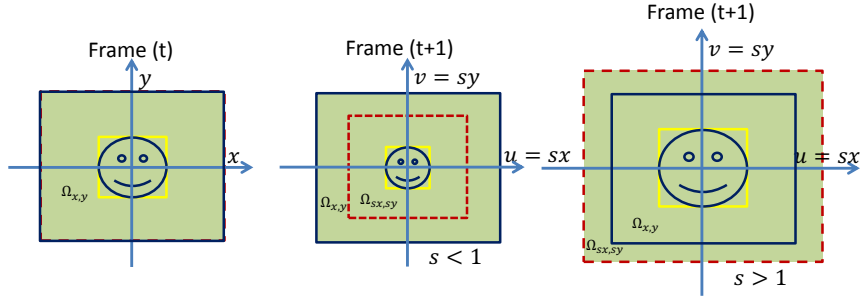}
\vspace{-2mm}
\end{center}
   \caption{Illustration of scale change. From left to right, the scale ratio is $s$. $\Omega_{x,y}$ inside the solid rectangles denotes the context region at the $t$-th frame, and its corresponding context region at the ($t+1$)-th frame is denoted by $\Omega_{sx,sy}$ that is inside the dotted rectangles.}
\label{fig:scale}
\end{figure}
\begin{figure}[t]
\begin{center}
\includegraphics[width=0.5\linewidth]{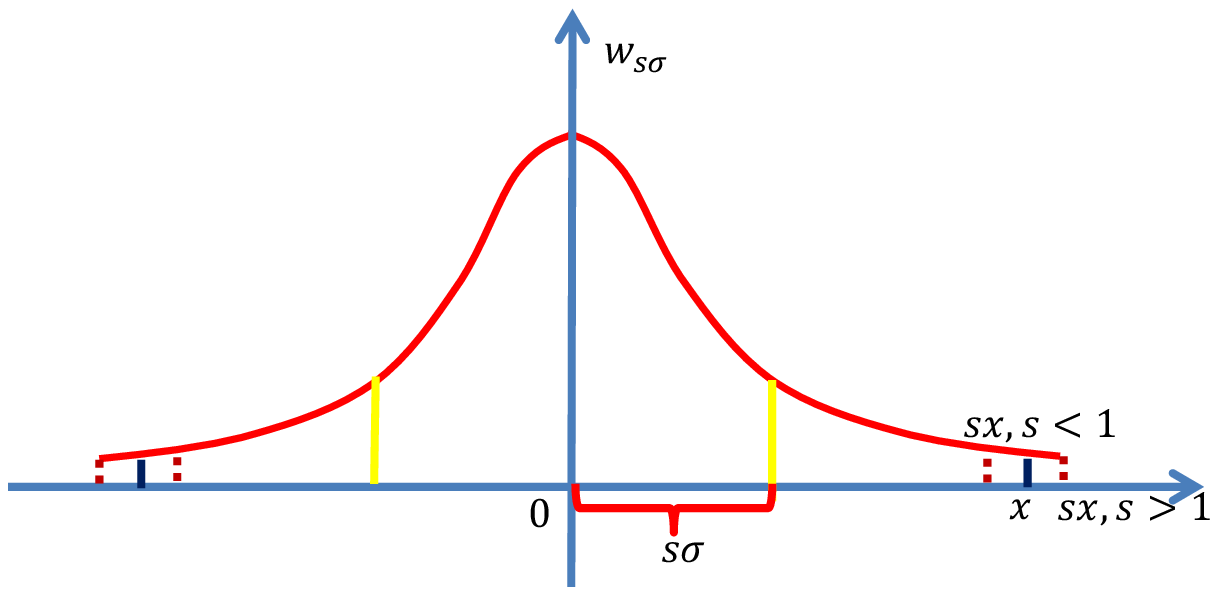}
\vspace{-2mm}
\end{center}
   \caption{Illustration of 1-D cross section of the weight function $w_{s\sigma}(\textbf{x})$.}
\label{fig:scaleweight}
\end{figure}
%
%

Without loss of generality, we assume the target object is centered at $\textbf{x}^\star=(0,0)$. Then, the confidence map (i.e.,~(\ref{eq:confmapdeftt})) can be represented as
\begin{equation}
c(\textbf{x})=H(\textbf{x})\otimes (I(\textbf{x})w_{\sigma}(\textbf{x})).
\label{eq:confmap}
\end{equation}
Then, we have
\begin{equation}
c(0,0)=\int\int_{\Omega_{x,y}}H(x,y)I(-x,-y)w_{\sigma}(-x,-y)dxdy.
\label{eq:confmap}
\end{equation}
See Figure~\ref{fig:scale}, when size of the target changes from left to right with ratio $s$, performing a change of variables $(u,v)=(sx,sy)$, we can reformulate (\ref{eq:confmap})
\begin{equation}
\begin{aligned}
c_t(0,0)&=\int\int_{\Omega_{x,y}}H_t(x,y)I_t(-x,-y)w_{\sigma_t}(-x,-y)dxdy\\
      &=\int\int_{\Omega_{sx,sy}}H_t(u/s,v/s)I_t(-u/s,-v/s)w_{\sigma_t}(-u/s,-v/s)\frac{1}{s^2}dudv\\
      &=\int\int_{\Omega_{sx,sy}}H_t(u/s,v/s)I_{t+1}(-u,-v)w_{\sigma_t}(-u/s,-v/s)\frac{1}{s^2}dudv\\
      &=\int\int_{\Omega_{sx,sy}}H_t(u/s,v/s)w_{s\sigma_t}(-u,-v)I_{t+1}(-u,-v)\frac{1}{s^2}dudv\\
      &\approx\int\int_{\Omega_{sx,sy}}H_{t+1}(u,v)w_{s\sigma_t}(-u,-v)I_{t+1}(-u,-v)\frac{1}{s^2}dudv\\
      &=\int\int_{\Omega_{x,y}}H_{t+1}(u,v)w_{s\sigma_t}(-u,-v)I_{t+1}(-u,-v)\frac{1}{s^2}dudv-\underbrace{\int\int_{\Omega_{x,y}\backslash\Omega_{sx,sy}}H_{t+1}(u,v)w_{s\sigma_t}(u,v)I_{t+1}(-u,-v)\frac{1}{s^2}dudv}_{\approx 0\ \textrm{because}\ w_{s\sigma_t}(-u,-v)\approx 0\ \textrm{for}\ \textrm{all}\ (u,v)\in \Omega_{x,y}\backslash\Omega_{sx,sy} (\mathrm{See~Figure}~\ref{fig:scaleweight})}\\
      &\approx\int\int_{\Omega_{x,y}}H_{t+1}(u,v)w_{s\sigma_t}(-u,-v)I_{t+1}(-u,-v)\frac{1}{s^2}dudv.\\ \label{eq:scale}
\end{aligned}
\end{equation}
In (\ref{eq:scale}), we have used the following relationships
\begin{equation}
H_t(u/s,v/s)\approx H_{t+1}(u,v),
\end{equation}
\begin{equation}
I_t(u/s,v/s)\approx I_{t+1}(u,v).\\
\end{equation}
Because of the proximity between two consecutive frames, as in~\cite{collins2003mean}, we can make the above reasonable assumptions which are spatially scaled versions of $I_t$ and $H_t$, respectively.

It is difficult to estimate $s$ from the (\ref{eq:scale}) because of the nonlinearity of the Gaussian weight function $w_{s\sigma_t}$. We adopt an iterative method to approximately obtain $s$. We utilize the estimated scale $s_t$ at frame $t$ to replace the scale term $s$ in the Gaussian window $w_{s\sigma_t}$, and the other scale term that needs to estimate is denoted as $s_{t+1}$. Thus, (\ref{eq:scale}) is reformulated as
\begin{equation}
\begin{aligned}
c_t(0,0)      &\approx\int\int_{\Omega_{x,y}}H_{t+1}(u,v)w_{s_t\sigma_t}(-u,-v)I_{t+1}(-u,-v)\frac{1}{s_{t+1}^2}dudv\\
      &=\int\int_{\Omega_{x,y}}H_{t+1}(u,v)w_{\sigma_{t+1}}(-u,-v)I_{t+1}(-u,-v)\frac{1}{s_{t+1}^2}dudv\\
      &= \frac{1}{s_{t+1}^2}c_{t+1}(0,0),
\label{eq:approxscale}
\end{aligned}
\end{equation}
where we denote
\begin{equation}
\sigma_{t+1}=s_t\sigma_t.
\end{equation}
Thus, we have
\begin{equation}
s_{t+1}=\sqrt{\frac{c_{t+1}(0,0)}{c_t(0,0)}}.
\end{equation}
We average the scales estimated from the former $n$ consecutive frames to make the current estimation more stable
\begin{equation}
\overline{s}_t=\frac{1}{n}\sum_{i=1}^{n} s_{t-i}=\frac{1}{n}\sum_{i=1}^{n}\sqrt{\frac{c_{t-i}(0,0)}{c_{t-i-1}(0,0)}}.
\label{eq:scale-s}
\end{equation}

To avoid oversenstive scale adaptation, we utilize the follow equation to incrementally update the estimated scale
\begin{equation}
s_{t+1}=(1-\lambda)s_t + \lambda\overline{s}_t.
\label{eq:scaleS}
\end{equation}
\bibliographystyle{ieeetr}
\bibliography{egbib}

\begin{thebibliography}{10}

\bibitem{Yilmaz_CSUR_2006}
A.~Yilmaz, O.~Javed, and M.~Shah, ``Object tracking: A survey,'' {\em ACM
  Computing Surveys}, vol.~38, no.~4, 2006.

\bibitem{collins2003mean}
R.~T. Collins, ``Mean-shift blob tracking through scale space,'' in {\em CVPR},
  vol.~2, pp.~II--234, 2003.

\bibitem{Collins_PAMI_2005}
R.~T. Collins, Y.~Liu, and M.~Leordeanu, ``Online selection of discriminative
  tracking features,'' {\em PAMI}, vol.~27, no.~10, pp.~1631--1643, 2005.

\bibitem{Adam_CVPR_2006}
A.~Adam, E.~Rivlin, and I.~Shimshoni, ``Robust fragments-based tracking using
  the integral histogram,'' in {\em CVPR}, pp.~798--805, 2006.

\bibitem{yang2007spatial}
M.~Yang, J.~Yuan, and Y.~Wu, ``Spatial selection for attentional visual
  tracking,'' in {\em CVPR}, pp.~1--8, 2007.

\bibitem{Ross_IJCV_2008}
D.~Ross, J.~Lim, R.~Lin, and M.-H. Yang, ``Incremental learning for robust
  visual tracking,'' {\em IJCV}, vol.~77, no.~1, pp.~125--141, 2008.

\bibitem{Kwon_CVPR_2010}
J.~Kwon and K.~M. Lee, ``Visual tracking decomposition.,'' in {\em CVPR},
  pp.~1269--1276, 2010.

\bibitem{kwon2011tracking}
J.~Kwon and K.~M. Lee, ``Tracking by sampling trackers,'' in {\em ICCV},
  pp.~1195--1202, 2011.

\bibitem{Mei_PAMI_2011}
X.~Mei and H.~Ling, ``Robust visual tracking and vehicle classification via
  sparse representation,'' {\em PAMI}, vol.~33, no.~11, pp.~2259--2272, 2011.

\bibitem{Li_CVPR_2011}
H.~Li, C.~Shen, and Q.~Shi, ``Real-time visual tracking using compressive
  sensing,'' in {\em CVPR}, pp.~1305--1312, 2011.

\bibitem{sevilla2012distribution}
L.~Sevilla-Lara and E.~Learned-Miller, ``Distribution fields for tracking,'' in
  {\em CVPR}, pp.~1910--1917, 2012.

\bibitem{oron2012locally}
S.~Oron, A.~Bar-Hillel, D.~Levi, and S.~Avidan, ``Locally orderless tracking,''
  in {\em CVPR}, pp.~1940--1947, 2012.

\bibitem{zhang2012robust}
T.~Zhang, B.~Ghanem, S.~Liu, and N.~Ahuja, ``Robust visual tracking via
  multi-task sparse learning,'' in {\em CVPR}, pp.~2042--2049, 2012.

\bibitem{Grabner_BMVC_2006}
H.~Grabner, M.~Grabner, and H.~Bischof, ``Real-time tracking via on-line
  boosting,'' in {\em BMVC}, pp.~47--56, 2006.

\bibitem{Kalal_CVPR_2010}
Z.~Kalal, J.~Matas, and K.~Mikolajczyk, ``Pn learning: Bootstrapping binary
  classifiers by structural constraints,'' in {\em CVPR}, pp.~49--56, 2010.

\bibitem{Hare_ICCV_2011}
S.~Hare, A.~Saffari, and P.~H. Torr, ``Struck: Structured output tracking with
  kernels,'' in {\em ICCV}, pp.~263--270, 2011.

\bibitem{Babenko_PAMI_2011}
B.~Babenko, M.-H. Yang, and S.~Belongie, ``Robust object tracking with online
  multiple instance learning,'' {\em PAMI}, vol.~33, no.~8, pp.~1619--1632,
  2011.

\bibitem{Zhang_ECCV_2012}
K.~Zhang, L.~Zhang, and M.-H. Yang, ``Real-time compressive tracking,'' in {\em
  ECCV}, pp.~864--877, 2012.

\bibitem{henriques2012circulant}
J.~Henriques, R.~Caseiro, P.~Martins, and J.~Batista, ``Exploiting the
  circulant structure of tracking-by-detection with kernels,'' in {\em ECCV},
  pp.~702--715, 2012.

\bibitem{divvala2009empirical}
S.~K. Divvala, D.~Hoiem, J.~H. Hays, A.~A. Efros, and M.~Hebert, ``An empirical
  study of context in object detection,'' in {\em CVPR}, pp.~1271--1278, 2009.

\bibitem{yang2009context}
M.~Yang, Y.~Wu, and G.~Hua, ``Context-aware visual tracking,'' {\em PAMI},
  vol.~31, no.~7, pp.~1195--1209, 2009.

\bibitem{grabner2010tracking}
H.~Grabner, J.~Matas, L.~Van~Gool, and P.~Cattin, ``Tracking the invisible:
  Learning where the object might be,'' in {\em CVPR}, pp.~1285--1292, 2010.

\bibitem{dinh2011context}
T.~B. Dinh, N.~Vo, and G.~Medioni, ``Context tracker: Exploring supporters and
  distracters in unconstrained environments,'' in {\em CVPR}, pp.~1177--1184,
  2011.

\bibitem{Wen_ECCV_2012}
L.~Wen, Z.~Cai, Z.~Lei, D.~Yi, and S.~Li, ``online spatio-temporal structure
  context learning for visual tracking,'' in {\em ECCV}, pp.~716--729, 2012.

\bibitem{torralba2003contextual}
A.~Torralba, ``Contextual priming for object detection,'' {\em IJCV}, vol.~53,
  no.~2, pp.~169--191, 2003.

\bibitem{belongie2002shape}
S.~Belongie, J.~Malik, and J.~Puzicha, ``Shape matching and object recognition
  using shape contexts,'' {\em PAMI}, vol.~24, no.~4, pp.~509--522, 2002.

\bibitem{wolf2006critical}
L.~Wolf and S.~Bileschi, ``A critical view of context,'' {\em IJCV}, vol.~69,
  no.~2, pp.~251--261, 2006.

\bibitem{oppenheim1983signals}
A.~V. Oppenheim, A.~S. Willsky, and S.~H. Nawab, {\em Signals and systems},
  vol.~2.
\newblock Prentice-Hall Englewood Cliffs, NJ, 1983.

\bibitem{bolme2009average}
D.~S. Bolme, B.~A. Draper, and J.~R. Beveridge, ``Average of synthetic exact
  filters,'' in {\em CVPR}, pp.~2105--2112, 2009.

\bibitem{bolme2010visual}
D.~S. Bolme, J.~R. Beveridge, B.~A. Draper, and Y.~M. Lui, ``Visual object
  tracking using adaptive correlation filters,'' in {\em CVPR}, pp.~2544--2550,
  2010.

\bibitem{Grabner_ECCV_2008}
H.~Grabner, C.~Leistner, and H.~Bischof, ``Semi-supervised on-line boosting for
  robust tracking,'' in {\em ECCV}, pp.~234--247, 2008.

\bibitem{cehovin2013robust}
L.~Cehovin, M.~Kristan, and A.~Leonardis, ``Robust visual tracking using an
  adaptive coupled-layer visual model,'' {\em PAMI}, vol.~35, no.~4,
  pp.~941--953, 2013.

\end{thebibliography}
\end{document}